\documentclass{article}

\usepackage{arxiv}

\usepackage{hyperref}       
\usepackage{url}            
\usepackage{booktabs}       
\usepackage{amsfonts}       
\usepackage{nicefrac}       
\usepackage{microtype}      
\usepackage{lipsum}		
\usepackage{graphicx}
\usepackage{tikz}
\usepackage{enumitem}
\usepackage{caption}
\usepackage{subcaption}
\usepackage{pifont}

\usepackage{placeins}
\setlist[itemize]{leftmargin=*}

\usepackage[numbers,sort&compress]{natbib}  
\usepackage{array} 
\usepackage{doi}
\usepackage{xcolor}
\usepackage{multirow}
\usepackage{float}

\usepackage{listings}
\newcommand{\xmark}{\textcolor{red}{\ding{55}}} 
\newcommand{\checkmarkemoji}{\textcolor{green}{\ding{51}}} 

\newcolumntype{P}[1]{>{\centering\arraybackslash}p{#1}} %

\newcommand{\datakit}{\texttt{gridfm-datakit}\raisebox{-0.6ex}{\includegraphics[height=1.1em]{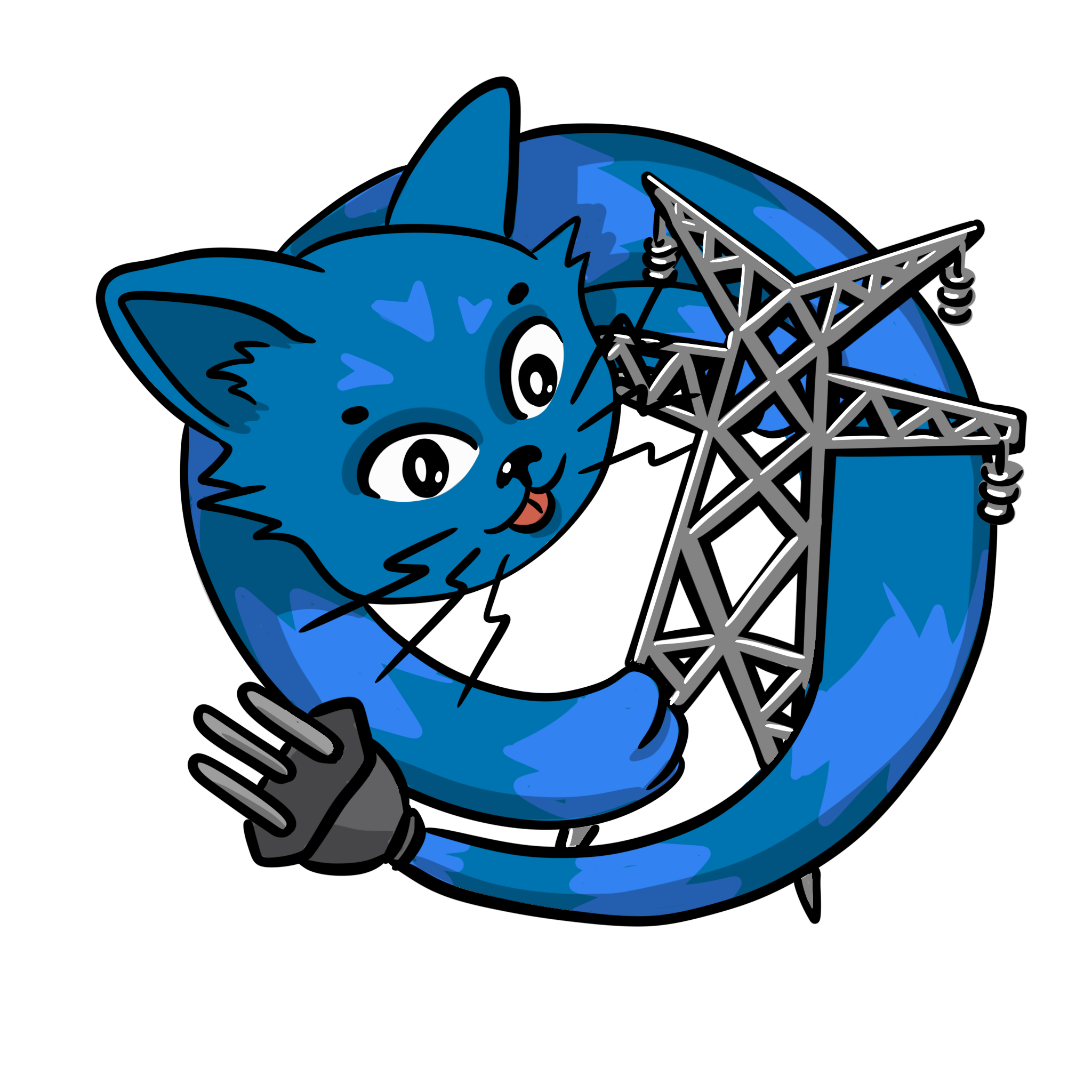}}\hspace{0.1em}}

\newcommand{\graphkit}{\texttt{gridfm-graphkit}\raisebox{-0.6ex}{\includegraphics[height=1.1em]{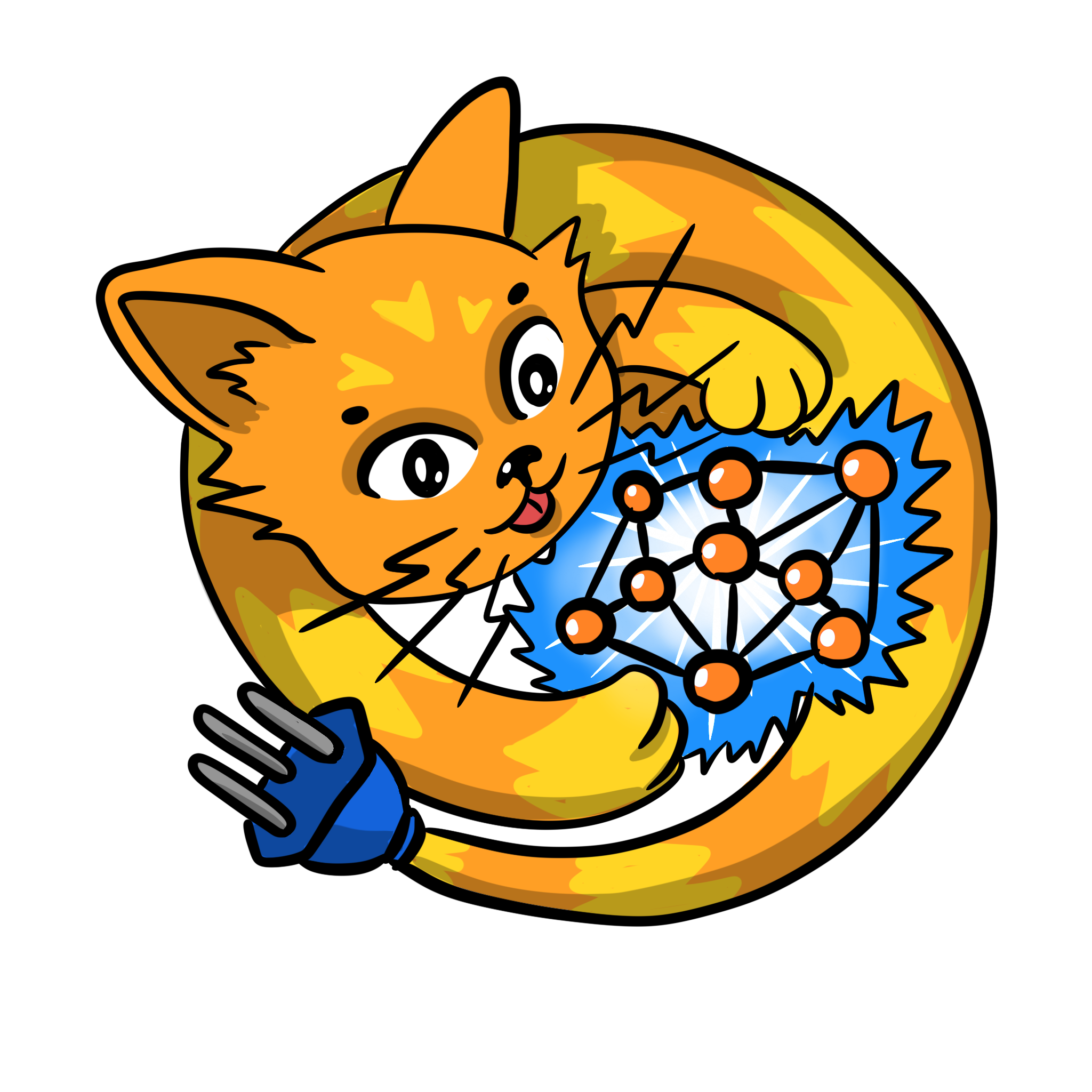}}\hspace{0.1em}}

\definecolor{keywordcolor}{RGB}{0,0,255}      
\definecolor{stringcolor}{RGB}{180,0,90}      
\definecolor{commentcolor}{RGB}{0,128,128}    
\definecolor{identifiercolor}{RGB}{0,0,0}     

\lstdefinestyle{pythonstyle}{
  language=Python,
  basicstyle=\ttfamily\footnotesize,
  keywordstyle=\color{keywordcolor}\bfseries,
  stringstyle=\color{stringcolor},
  commentstyle=\color{commentcolor}\itshape,
  identifierstyle=\color{identifiercolor},
  showstringspaces=false,
  breaklines=true,
  columns=fullflexible
}

\title{
    gridfm-datakit-v1: A Python Library for Scalable and Realistic Power Flow and Optimal Power Flow Data Generation}

\author{
\begin{tabular}{c}
Alban Puech$^{1,2}$\thanks{These authors contributed equally as main contributors.},
Matteo Mazzonelli$^{1}$\footnotemark[1],
Celia Cintas$^{1}$\thanks{Equal contributions.},
Tamara R. Govindasamy$^{1}$\footnotemark[2],\\
Mangaliso Mngomezulu$^{1}$\footnotemark[2],
Jonas Weiss$^{1}$\footnotemark[2],
Matteo Baù$^{3}$,
Anna Varbella$^{2}$,\\
François Mirallès$^{4}$,
Kibaek Kim$^{5}$,
Le Xie$^{6}$,
Hendrik F. Hamann$^{7,8}$,\\
Etienne Vos$^{1}$,
Thomas Brunschwiler$^{1}$
\end{tabular}
\\[6pt]
$^{1}$IBM Research \quad
$^{2}$ETH Zurich \quad
$^{3}$RSE S.p.A.\\
$^{4}$Hydro-Québec Research Institute \quad
$^{5}$Argonne National Laboratory\\
$^{6}$Harvard University \quad
$^{7}$Stony Brook University \quad 
$^{8}$Brookhaven National Laboratory
}

\renewcommand{\shorttitle}{\datakit-v1}

\begin{document}
\maketitle

\vspace{-1.2cm}
\begin{figure}[H]
  \centering

\includegraphics[width=0.75\textwidth]{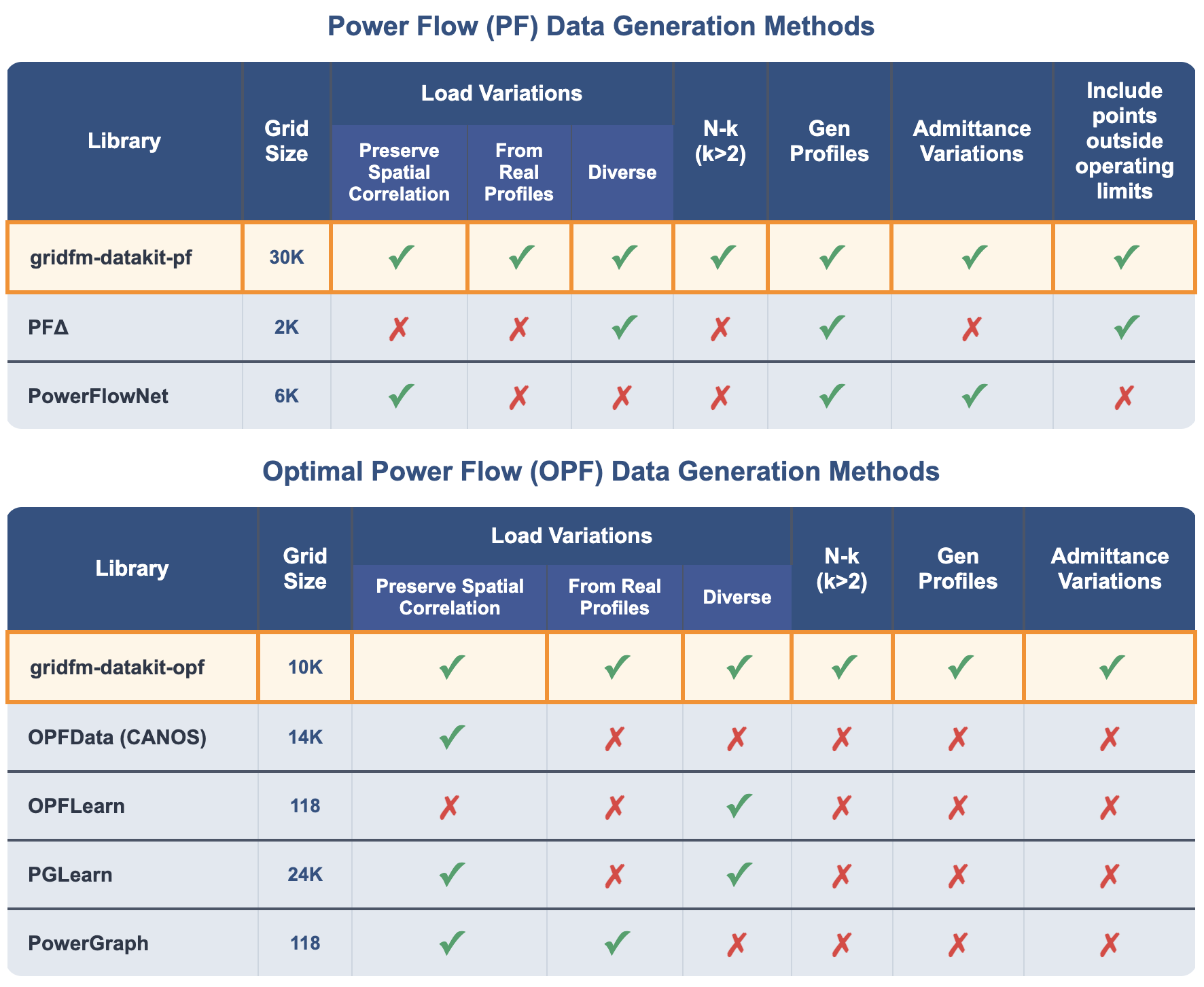}
  \caption{\datakit unifies and advances state-of-the-art methods for data generation that were previously scattered across existing libraries, providing the most scalable Python-based framework for generating both Power Flow and Optimal Power Flow datasets for the transmission grid.}
  \label{fig:summary}
\end{figure}

\newpage

\begin{abstract}
    \begin{itemize}
        \item Existing libraries for generating synthetic data for training ML-based PF and OPF solvers do not integrate all state-of-the-art techniques for producing diverse load, topology, and generator scenarios (see Figure~\ref{fig:summary}), which limits the performance of ML-based solvers.
        \item PF data generation is often restricted to OPF-feasible points, hindering generalization to cases that violate operating limits (e.g., with branch overloads or voltage violations).
        \item OPF datasets have low diversity due to fixed generator cost functions, limiting generalization across varying costs.
        \item We introduce \datakit, a Python library unifying stochastic yet realistic load and topology perturbations, arbitrary $N\!-\!k$ changes, and scaling to large grids (10,000 buses) to generate PF and OPF datasets.
        \item By combining global load scaling from real-world profiles with localized noise, it generates data with realism and diversity beyond prior libraries.
        \item We compare \datakit’s data distributions with those of \textsc{OPFData}, \textsc{OPF-Learn}, \textsc{PGLearn}, and \textsc{PF$\Delta$} in Section~\ref{sec:diversity}.
        \item It is available on \href{https://github.com/gridfm/gridfm-datakit}{\underline{GitHub}} under the Apache 2.0 license and can be installed via \texttt{pip install gridfm-datakit}.
    \end{itemize}
\end{abstract}

\vspace{1cm}

\section{Getting Started}

\datakit can be installed from \underline{\href{https://pypi.org/project/gridfm-datakit/}{PyPI}} and used interactively or via the command line. Full documentation is available \href{https://gridfm.github.io/gridfm-datakit/}{here}. Users are kindly requested to cite this technical paper when publishing work that uses \datakit.

\paragraph{Interactive Interface.}

The interactive interface guides the user through the different settings for data generation. It can be launched directly in a Jupyter Notebook:

\begin{lstlisting}[style=pythonstyle]
from gridfm_datakit.interactive import interactive_interface
interactive_interface()
\end{lstlisting}

\paragraph{Command-Line Interface.}

For automated or large-scale runs, the data generation routine can be executed from the following bash command line using a \underline{\href{https://github.com/gridfm/gridfm-datakit/tree/main/scripts/config}{configuration file}}:

\begin{lstlisting}[language=bash]
gridfm-datakit generate path/to/config.yaml
\end{lstlisting}

\section{Motivations}

The absence of standardized, realistic, and diverse datasets for electric grids hinders the development of machine learning (ML) methods for steady-state transmission grid analysis~\cite{pglearn}, motivating the design of \datakit.

\begin{itemize}
    \item \textbf{Barrier to entry.} Generating large datasets is computationally expensive. Solving millions of OPF problems can require hundreds of CPU hours\footnote{ \scriptsize For instance, \textsc{OPFData}~\cite{lovett2024opfdatalargescaledatasetsac} was created by solving 600{,}000 OPF cases for a 10k-bus grid using \texttt{PowerModels.jl}~\cite{powermodels}; we measured a convergence time of 17\,s per case, implying over 118 CPU-days of computation for this grid alone. As another example, generating the PGLearn dataset took 29,876.65 CPU-hours, i.e., \textbf{about 1245 CPU-days!}~\cite{pglearn}}.
    \item \textbf{Focus on small grids.} Most ML models have been evaluated only on networks with fewer than 1,000 buses \cite{DONON2020106547, varbella2024powergraph, LOPEZGARCIA2023105567, matteothesis}.
    While recent efforts have scaled to larger systems \cite{arowolo2025generalizationgraphneuralnetworks, piloto2024canosfastscalableneural,lovett2024opfdatalargescaledatasetsac, LIN2024110112, matteothesis, pfdelta}, they remain rare due to the high cost of solving OPF problems on large instances.
\item \textbf{Lack of reproducibility and benchmarking.} Most studies rely on custom, unpublished data pipelines with differing assumptions on loads, dispatch, and topology~\cite{DONON2020106547, varbella2024powergraph, LOPEZGARCIA2023105567}, hindering reproducibility and fair comparison. Limited incentives to generate diverse or complex scenarios also make it unclear whether reported gains arise from model design or data choices. \datakit provides a unified platform for producing larger datasets to support emerging standardized benchmarks for PF solvers, such as those in \textsc{PF$\Delta$}~\cite{pfdelta}\footnote{ \scriptsize The authors welcome collaboration to extend~\cite{pfdelta} with \datakit datasets for larger grids and for OPF.}.
\end{itemize}

As a result, progress on critical challenges (such as transferability and generalization to unseen grid states), as discussed in a recent review paper~\cite{Khaloie}, is, for now, limited.

\section{Limitations of Current Libraries}

Existing libraries show several limitations that reduce their usability:

\begin{itemize}
    \item \textbf{Incomplete perturbation modeling.}  
State-of-the-art methods for generating diverse and realistic load, topology, generator dispatch, and admittance scenarios are scattered across more than ten Python and Julia libraries, and even the latest ones don't include all of them, as shown in Figure~\ref{fig:summary}. In particular, most datasets only consider $N-1$ contingencies, although real grids can see ten or more branches switch status per day. 

    \item \textbf{PF datasets are limited to points within operating limits.}  
    Most PF libraries only generate points that are feasible for the AC Optimal Power Flow (ACOPF) problem, excluding operating states that violate its inequality constraints (e.g., voltage magnitude or branch limits). This prevents using such datasets to train PF solvers that need to handle cases that violate these constraints. \textsc{PF$\Delta$} solves this issue by employing an OPF formulation that neglects operating limits, but this leads to clusters of points showing the same violations on a few buses, as discussed in Section~\ref{sec:diversity}.

    \item \textbf{OPF datasets use fixed generator costs.}  
    OPF libraries (e.g., \textsc{OPFData}~\cite{lovett2024opfdatalargescaledatasetsac}, \textsc{PGLearn}~\cite{pglearn}, \textsc{OPFLearn}~\cite{opflearn}) use fixed generator cost coefficients, limiting the diversity of generator dispatch and the ability for models trained on these datasets to generalize across different cost conditions.
\end{itemize}

We additionally provide a detailed comparison of all existing libraries in Appendix~\ref{app:comparison}. In what follows, we show how \datakit overcomes these limitations to generate OPF data, as well as PF data that can include points violating the OPF inequality constraints.

\section{Design and workflow}

\subsection{Grid, Size, and Scaling}
\datakit scales to networks with up to 30,000 buses for PF and 10,000 buses for OPF using \texttt{PowerModels.jl}~\cite{powermodels}. Grids can be imported from the MATPOWER format (\texttt{.m})~\cite{matpower}, and all grids from the \textsc{PGLib} dataset~\cite{pglib} are supported. This makes our library the most scalable Python-based solution for AC PF and OPF data generation. As shown in Table~\ref{tab:scalability}, \textbf{data generation is fast}: generating 200,000 PF samples for case24 and case118 took \textbf{less than 10 minutes and 20 minutes}, respectively. For OPF data, this required about \textbf{1 hour and 2 hours}, respectively, using only 20 cores and 32 GB of RAM.

\begin{table}[t]
\centering
\caption{Time required to generate approximately 200,000 data points with $n$-1 topology perturbations for different grid sizes. See Appendix~\ref{app:data_gen} for details on data generation runtime.\newline}
\label{tab:scalability}
\resizebox{0.9\linewidth}{!}{%
\begin{tabular}{lccc}
\hline
Grid name    & Number of samples obtained & CPU hours & Convergence rate (\%) \\
\hline 
\multicolumn{4}{c}{\textbf{Power Flow (PF)}} \\
\hline
IEEE 24-bus  & 199,540  & 2.69      & 99.77     \\
IEEE 118-bus & 199,339  & 6.44      & 99.67     \\
GOC 2,000     & 198,858  & 247.55    & 99.43     \\
GOC 10,000      & 199,880  & 1,384.24  & 99.94     \\
\hline
\multicolumn{4}{c}{\textbf{Optimal Power Flow (OPF)}} \\
\hline
IEEE 24-bus  & 190,387  & 21.33     & 95.19     \\
IEEE 118-bus & 197,769  & 46.10     & 98.88     \\
GOC 2,000     & 198,308  & 1,103.67  & 99.15     \\
GOC 10,000      & 195,920  & 3,628.01  & 97.96     \\
\hline
\end{tabular}}
\end{table}

\subsection{Load Scenarios}

Most (O)PF libraries and papers (e.g., \textsc{Graph Neural Solver}~\cite{DONON2020106547}, \textsc{OPFData}~\cite{lovett2024opfdatalargescaledatasetsac}, \textsc{PowerFlowNet}~\cite{LIN2024110112}) apply uncorrelated synthetic load perturbations, typically scaling each nominal bus load independently with uniform or Gaussian noise, leading to very little load diversity. \textsc{PGLearn}~\cite{pglearn} introduces a global scaling factor applied to all buses, with additional local bus-level noise, yet both are sampled from uniform distributions. \textsc{PowerGraph}~\cite{varbella2024powergraph} leverages real aggregated load profiles to derive time-dependent global scaling factors, but omits local noise, resulting in limited spatial diversity. Finally, \textsc{OPF-Learn}~\cite{opflearn} and \textsc{PF$\Delta$}~\cite{pfdelta} sample directly from the feasible load space, producing scenarios that may be unrealistically uncorrelated across scenarios and buses, and computationally expensive to generate for large networks\footnote{\scriptsize This sampling technique prevents \textsc{PF$\Delta$} and \textsc{OPF-Learn} from generating data for grids larger than 2{,}000 buses and from producing large datasets for 2{,}000-bus grids.}.

We introduce a \textbf{hybrid load perturbation strategy} that combines global scaling from real aggregated time series (e.g., aggregated load profiles obtained from the EIA~\cite{EIA_ERCOT_2025}) with local multiplicative noise at each bus. Let $p_i$ and $q_i$ denote the nominal active and reactive powers of load $i$. At each time step $t$, a global scaling factor $\texttt{ref}_t$ is derived from an aggregated profile scaled into a feasible range $[l, u]$\footnote{ \scriptsize $u$ is obtained by incrementally increasing the nominal active and reactive power loads in 10\% steps until OPF no longer converges. $l$ is then set to $(1-r)u$, where $r$ is a parameter defaulting to $0.4$, as in \textsc{PGLearn}~\cite{pglearn}.}. The perturbed loads are then given by
\[
\tilde{p}_{i,t} = p_i \cdot \texttt{ref}_t \cdot \epsilon^p_{i,t}, \quad
\tilde{q}_{i,t} = q_i \cdot \texttt{ref}_t \cdot \epsilon^q_{i,t}
\]
where $\epsilon^p_{i,t}, \epsilon^q_{i,t} \sim \mathcal{U}(1 - \sigma, 1 + \sigma)$ introduce per-bus variation.

This approach jointly preserves \textit{spatial correlation and realism} (via shared global scaling factors), captures \textit{temporal realism} (via real load profiles), and ensures \textit{spatial diversity} (via local noise), as shown in Figure~\ref{fig:summary}.

\subsection{Topology and Admittance Perturbations}

While libraries are restricted to $N-1$ contingencies, i.e., single-line, transformer, or generator outages~\cite{pfdelta, opflearn, lovett2024opfdatalargescaledatasetsac}, \textbf{datakit supports arbitrary $N-k$ perturbations}. This can be done in two different ways: by exhaustively enumerating all valid topologies with up to $k$ disconnected components (lines/transformers/generators), or by randomly sampling topologies by disabling up to $k$ components, while ensuring feasibility (i.e., no islanding).

We introduce admittance perturbations by randomly scaling the resistance and reactance of branches using scaling factors sampled from a uniform distribution in the range \(( \max(0, 1 - \sigma), 1 + \sigma )\), where \(\sigma\) is a user-defined parameter.

\subsection{Generator Setpoints}

We provide two data modes: one intended for generating datapoints for training OPF solvers \textbf{(OPF mode)}, with cost-optimal dispatches that satisfy all operating limits (OPF-feasible), and one intended for training PF solvers \textbf{(PF mode)}, which produces power flow data where one or more operating limits -- the inequality constraints defined in OPF, e.g., voltage magnitude or branch limits  -- may be violated.

\paragraph{OPF mode.}

Generator setpoints are obtained by solving an ACOPF problem after topology perturbations. The resulting operating points satisfy all limits and are cost-optimal for the chosen generator cost coefficients.

Unlike typical OPF data libraries that use fixed generator costs~\cite{opflearn, pglearn}, \texttt{gridfm-datakit} increases dispatch diversity by permuting cost coefficients or applying user-defined random scaling factors before solving the OPF. This supports training models that generalize across different cost or market conditions.

\paragraph{PF mode.}
Several strategies have been proposed in the literature to generate PF data that include points outside normal operating limits. \textsc{PowerFlowNet}~\cite{LIN2024110112}, \textsc{TypedGNN}~\cite{LOPEZGARCIA2023105567}, and \textsc{Graph Neural Solver}~\cite{DONON2020106547} directly sample voltage magnitudes and generator powers within their admissible ranges. However, this approach restricts scenarios to normal operating limits, prevents the use of realistic load scenarios (since the load is then set according to the sampled dispatch), and results in similar total system load across samples. \textsc{PF$\Delta$}~\cite{pfdelta} instead removes inequality constraints during OPF solving, but this causes most violations to cluster in a few locations, as shown in Figure~\ref{fig:qg}.

In \texttt{gridfm-datakit}, generator setpoints are first computed on the \textit{base topology} (with load and admittance perturbations) using ACOPF. After fixing these setpoints, topology perturbations are applied, and an AC power flow is solved to obtain the new system state. Because generator dispatch is not re-optimized after the topology change, some samples violate operating limits defined by the OPF inequality constraints, such as branch overloads, voltage limit violations, branch angle difference violations, reactive power bound violations, and active power bound violations at the slack bus.

These cases emerge naturally from realistic perturbations rather than from artificial constraint removal or direct sampling. This approach provides a \textbf{balanced mix of points within and outside normal operating limits}, reflecting \textbf{realistic system behavior} where OPF determines generator dispatch under nominal conditions and unexpected changes lead to violations. 

\section{Data Diversity} \label{sec:diversity}

\begin{figure}[H]
    \centering
    \begin{subfigure}[t]{0.48\textwidth}
        \centering
        \includegraphics[width=\linewidth]{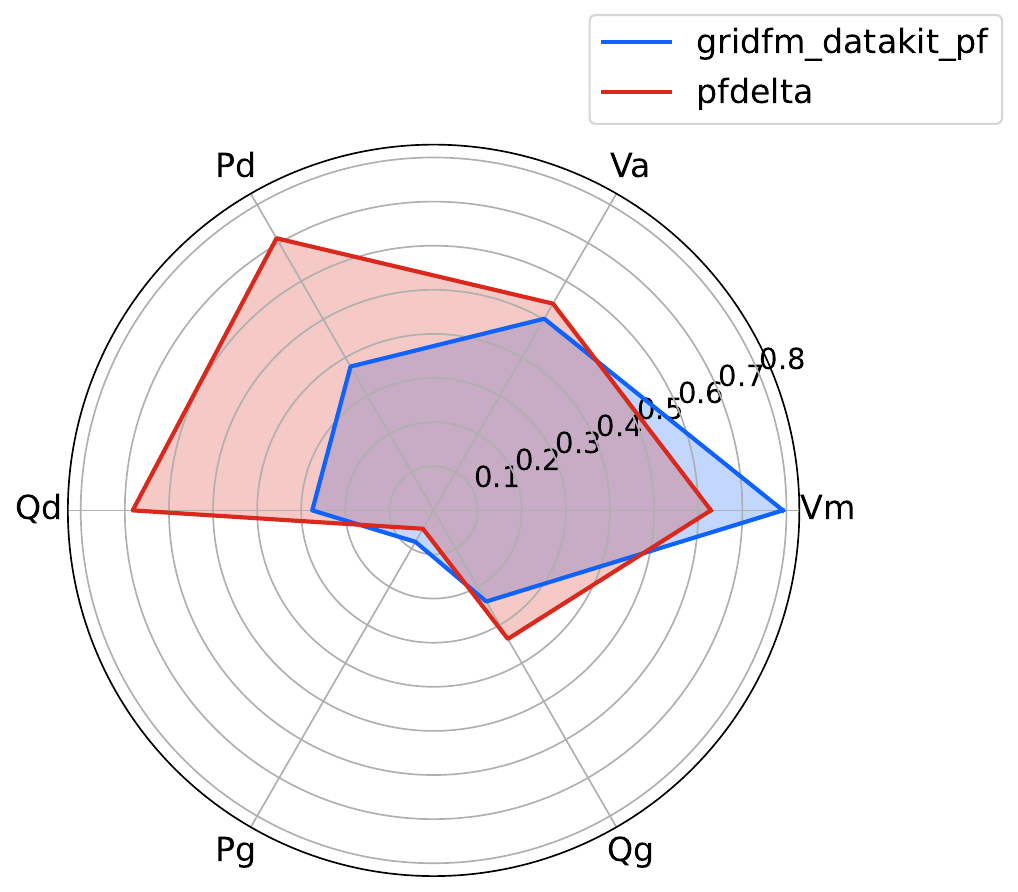}
        \caption{Power Flow (PF) Datasets}
        \label{fig:spider_pf}
    \end{subfigure}
    \hfill
    \begin{subfigure}[t]{0.48\textwidth}
        \centering
        \includegraphics[width=\linewidth]{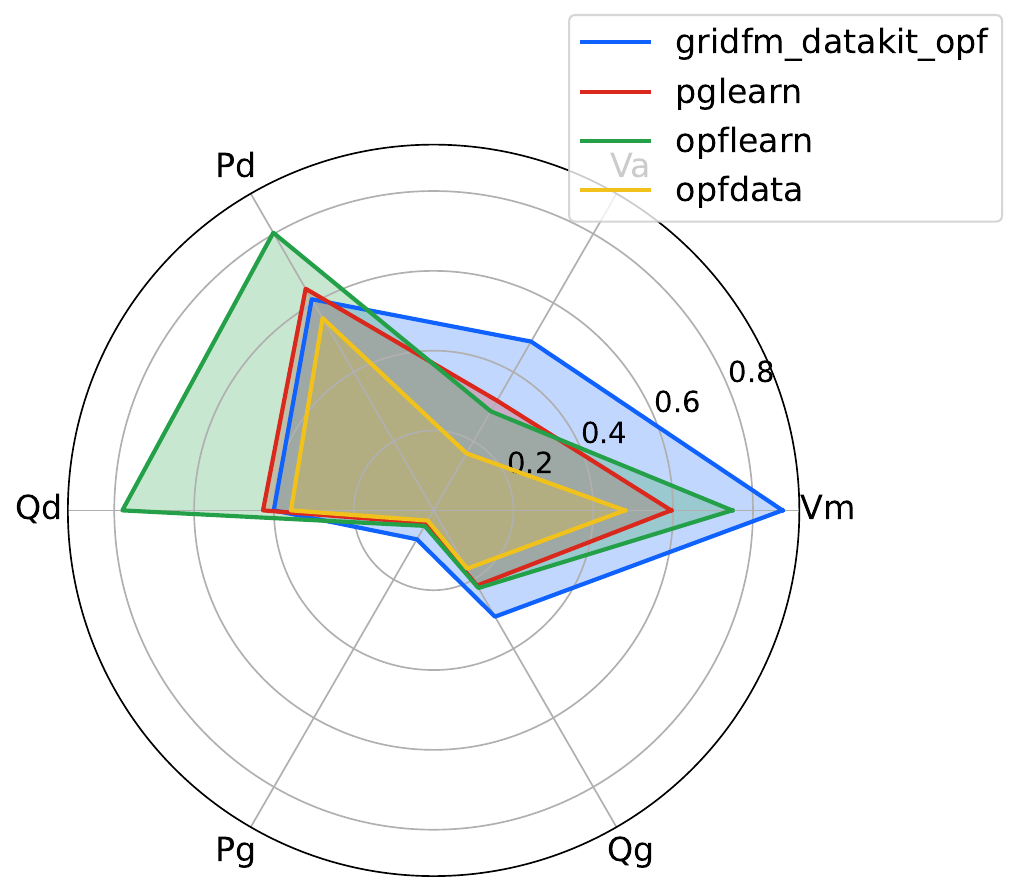}
        \caption{Optimal Power Flow (OPF) Datasets}
        \label{fig:spider_opf}
    \end{subfigure}
    \caption{Normalized mean feature entropy across PF (left) and OPF (right) datasets.}
    \label{fig:spider}
\end{figure}

We assess the diversity of datasets generated with \datakit against existing PF and OPF libraries\footnote{  \scriptsize To ensure a fair comparison with other libraries, we remove at most one line, transformer, or generator in each sample (as in all other libraries except \textsc{OPF-Learn}). The parameter $\sigma$ of the local load perturbations is set to 0.2 (as in \textsc{PGLearn}, and \textsc{OPFData}), and the range parameter of the global scaling factor is set to 0.4 (as in \textsc{PGLearn}). Generator costs are permuted as in \textsc{PF$\Delta$}, and the $\sigma$ parameter of the admittance perturbation is set to 0.2 (a feature absent in other libraries).}. Figure~\ref{fig:spider} shows, for each feature, the mean normalized Shannon entropy, an information-theoretic measure analogous to metric $Q_1$ in \cite{hedgeopf2025}. More details on the computation of this metric are provided in Appendix~\ref{app:shannon}.

\paragraph{Power flow data.}

Among recent and publicly available PF libraries, only \textsc{PF$\Delta$} is suitable for comparison, as it is the only one that includes out-of-operating-limit scenarios (contrary to, e.g., \textsc{PowerFlowNet}). \datakit is less diverse for loads ($P_d$, $Q_d$) because \textsc{PF$\Delta$} samples the feasible load space directly, whereas \datakit applies global load scaling from real profiles plus local noise, balancing spatial and temporal correlation with diversity. Uniform sampling yields high variability in power factors across buses and samples. In contrast, \datakit perturbs power factors only through local noise, while the global scaling jointly applied to active and reactive loads prevents large variations.

\begin{figure}[H]
    \centering
        \begin{minipage}{0.48\linewidth}
        \centering
        \includegraphics[width=\linewidth]{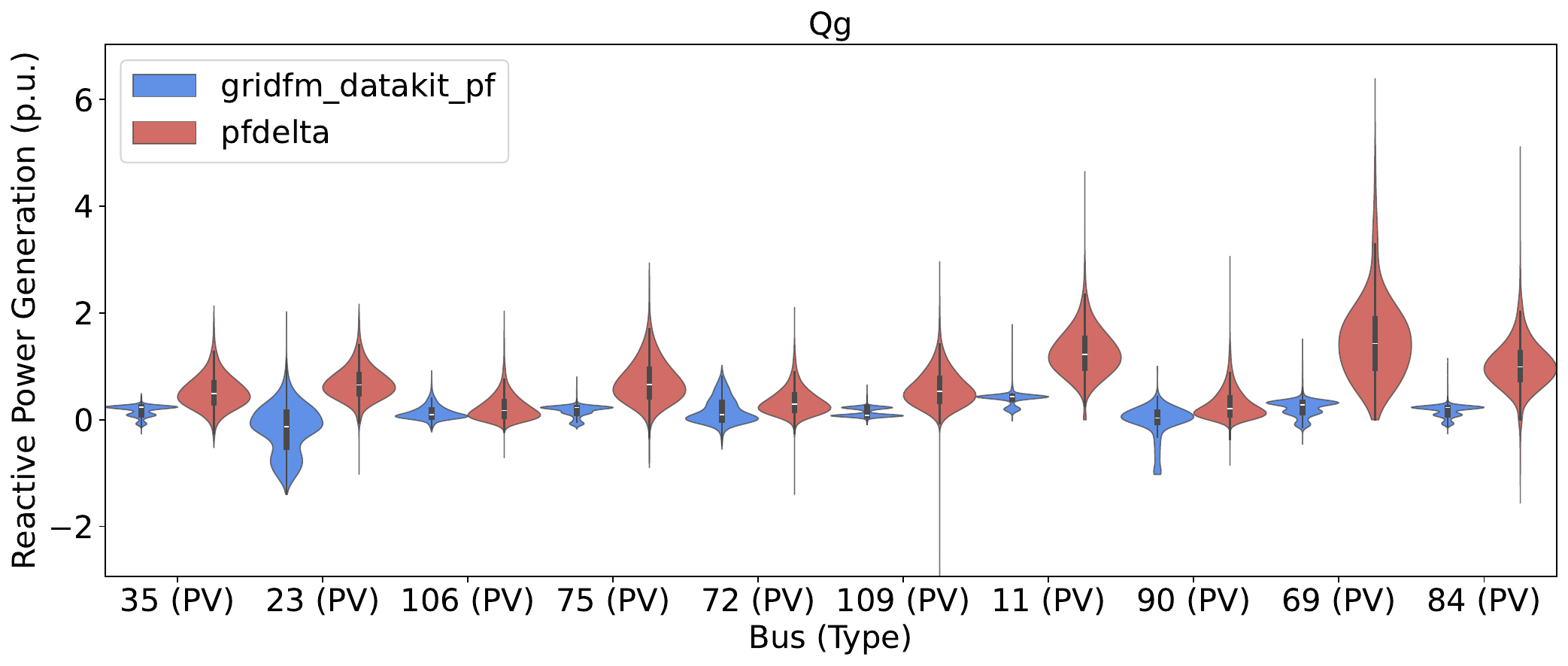}
        \caption{Distribution of reactive power generation ($Q_g$) for PF datasets (\datakit\ vs.~\textsc{PF$\Delta$}). More than 75\% of the samples in \textsc{PF$\Delta$} violate the reactive power bounds at buses~69 and~84, which are 0.32~p.u. and 0.23~p.u., respectively.}
        \label{fig:qg}
    \end{minipage}\hfill
    \begin{minipage}{0.48\linewidth}
        \centering
        \includegraphics[width=\linewidth]{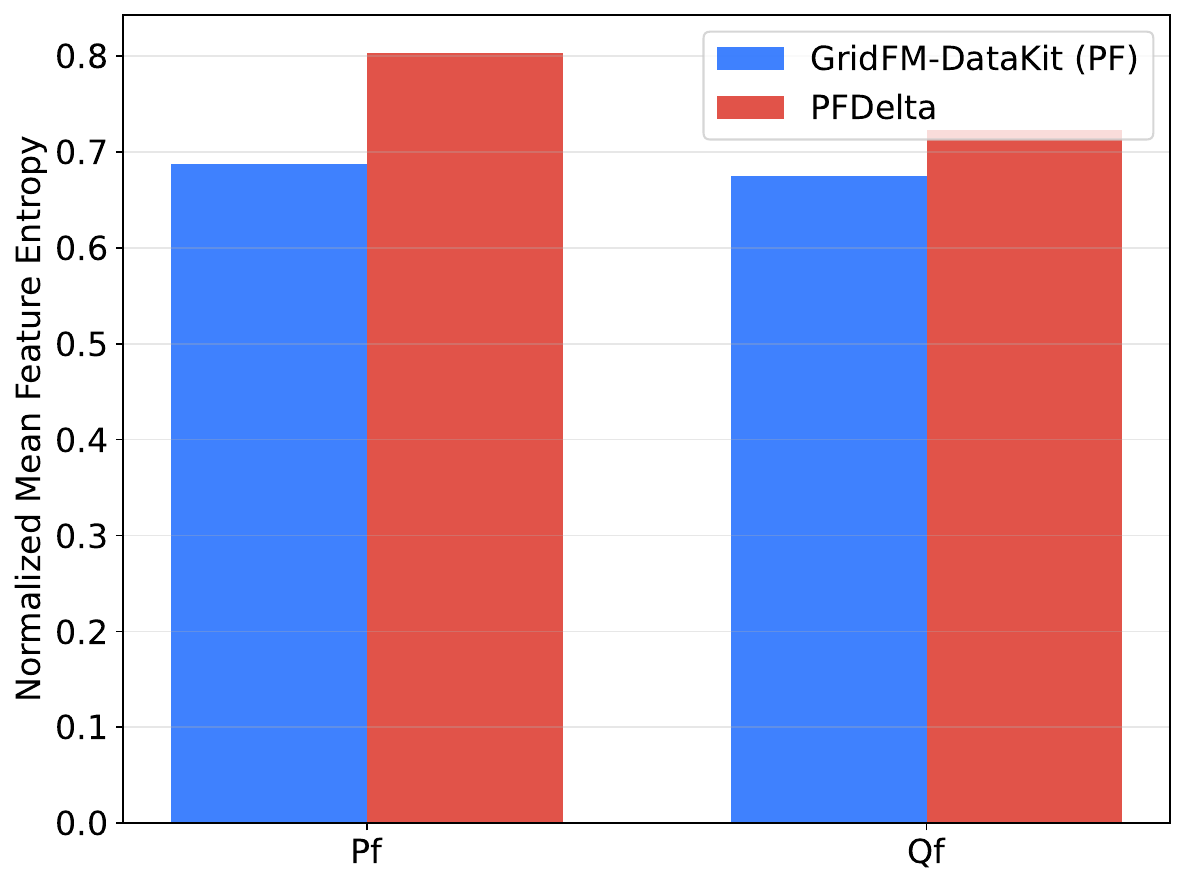}
        \caption{Branch flow entropy ($P_f$, $Q_f$).}
        \label{fig:barplot}
    \end{minipage}

\end{figure}

\datakit exhibits slightly lower diversity than \textsc{PF$\Delta$} for $Q_g$ -- expected since \textsc{PF$\Delta$} solves OPF without inequality constraints, which naturally leads to a higher proportion of samples outside normal operating limits. However, this higher diversity in \textsc{PF$\Delta$} comes at the cost of realism and balance between points within and outside nominal operating limits (e.g., more than 75\% of samples violate reactive power limits at buses~69 and~84 in Figure~\ref{fig:qg}).

\begin{figure}[H]
    \centering
    \begin{subfigure}[t]{0.48\linewidth}
        \centering
        \includegraphics[width=\linewidth]{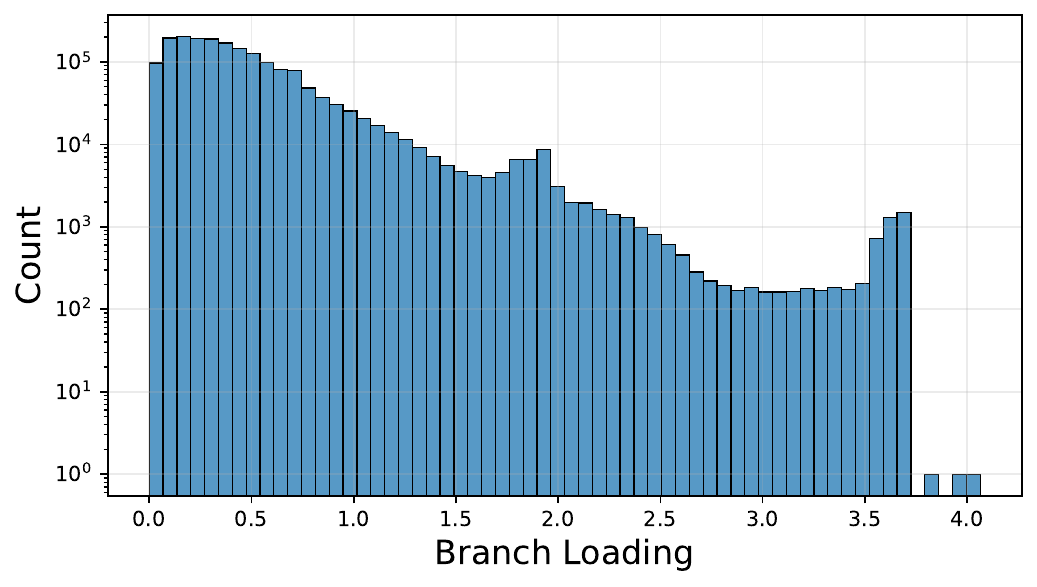}
        \caption{\textsc{PF$\Delta$}}
        \label{fig:pfdelta_hist}
    \end{subfigure}
    \hfill
    \begin{subfigure}[t]{0.48\linewidth}
        \centering
        \includegraphics[width=\linewidth]{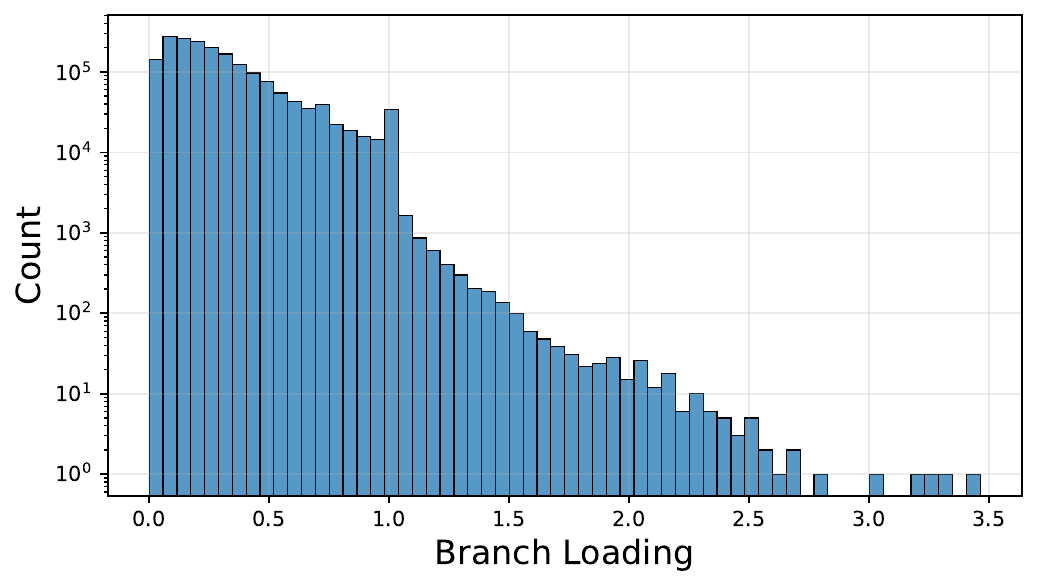}
        \caption{\datakit}
        \label{fig:loading_hist_gridfm}
    \end{subfigure}
    \caption{Distribution of branch loading for PF datasets (log-scale y-axis).}
    \label{fig:loading_hist}
\end{figure}

We show in Figure~\ref{fig:barplot} the mean normalized entropy for branch flow features and in Figure~\ref{fig:loading_hist} the distribution of branch loadings for both datasets. Both show similar entropy, but \datakit and \textsc{PF$\Delta$} differ in the balance between out-of-limit and in-limit scenarios: in \datakit, $1.2\%$ of the branches are overloaded (loading $>$ 1) across all scenarios, and $79\%$ of the scenarios have at least one branch overloading. In \textsc{PF$\Delta$}, all scenarios have overloads (as all inequality constraints are removed), and $8\%$ of all branches are overloaded.

\paragraph{Optimal power flow data.}

We compare \datakit with \textsc{OPFData}, \textsc{OPF-Learn} and \textsc{PGLearn}. \datakit shows similar diversity to \textsc{OPFData} and \textsc{PGLearn} for load features ($P_d$, $Q_d$), though with more realistic load distributions due to the hybrid load perturbation strategy. \textsc{OPF-Learn} shows higher diversity for load, since it samples directly from the feasible load space (leading to the aforementioned issues of unrealistic power factor variations, and lack of spatial and temporal correlation). For all other features, \datakit\ demonstrates substantially greater diversity, primarily driven by the permutation of generator cost coefficients and the use of admittance perturbations, which are absent in other datasets. Notably, \datakit achieves a normalized mean entropy for $P_g$ that is 2.3 times higher than \textsc{PGLearn} and \textbf{2.8 times higher than \textsc{OPFData}}, although all datasets still struggle to achieve high diversity for $P_g$, which is the main decision variable in OPF.

Finally, Figure~\ref{fig:feature_violin_type} in Appendix~\ref{app:violin} compares the feature distributions for randomly selected buses across datasets for 10{,}000 samples, following the methodology of~\cite{pfdelta}.

\subsection{Outputs, Data Validation, and Benchmarking}

\paragraph{Outputs.}
\datakit produces structured outputs suitable for direct use in machine learning pipelines. Each generated scenario includes complete per-bus, per-branch, and per-generator data, as shown in Figure~\ref{fig:outputs} in Appendix~\ref{app:outputs}.

\paragraph{Benchmarking.}

Fast linear models such as DC-PF and DC-OPF often serve as baselines for neural PF/OPF solvers. To avoid re-running external solvers, \datakit computes DC-PF and DC-OPF for every generated sample and stores the solutions and the per-sample runtime directly in the dataset.

\paragraph{Data validation and statistics.}

The CLI command \texttt{gridfm\_datakit validate path/to/dataset} checks all grid-model assumptions, recomputes constraint satisfaction, and verifies consistency between computed quantities (flows, balance, loading) and those stored in the dataset, providing a reference implementation of these core calculations.

\texttt{gridfm\_datakit stats path/to/dataset} produces a statistical report with distributions of AC/DC power-balance residuals, runtime, branch loading, overload counts, and element counts, as shown in Figure~\ref{fig:stats} in Appendix~\ref{app:outputs}.

\section{From data to foundation models using \graphkit}

\graphkit~\cite{Graphkit} is a Python library designed as part of the GridFM project~\cite{hamann2024perspectivefoundationmodelselectric} to enable training, fine-tuning, and deployment of foundation models for electric power grids using datasets generated with \datakit. It provides seamless conversion of power grid data into \texttt{PyTorch Geometric}~\cite{PyG} data objects, where buses are represented as nodes and transmission lines as edges.

Built on \texttt{PyTorch Lightning} and \texttt{PyTorch Geometric}, \graphkit\ offers a modular framework for training Graph Neural Networks and Graph Transformers. It supports self-supervised pre-training via masked feature reconstruction combined with a physics-informed loss enforcing AC power balance equations.

The library is optimized for large-scale, multi-GPU training and supports zero-shot evaluation on unseen topologies, as well as fine-tuning for downstream tasks such as power flow or contingency analysis. Post-processing utilities compute line loadings, detect thermal limit violations, and identify critical components under perturbed conditions.

Through its modular design, \graphkit\ can be easily extended with new models, loss functions, and learning tasks, providing a unified environment for developing graph-based foundation models for the electric grid.

\section*{Limitations and Future Work}

Planned features include:
\begin{itemize}[leftmargin=*]
\item Support for topology variations in PF prior to defining the generator setpoints (using OPF), to increase the diversity of operating points without raising the proportion of constraint violations.
\item Support for bus-level load profiles, in addition to system-level aggregated ones (e.g., from ENTSO-E).
\item Sampling of subgraphs from existing grids to generate more diverse topologies.
\item Open-sourcing of large scale synthetic dataset produced by \datakit on \underline{\href{https://huggingface.co/gridfm}{HuggingFace GridFM}}. 
\item Saving dual solutions in the dataset for primal-dual learning~\cite{primaldual} or LMP prediction.
\end{itemize}

\section*{Acknowledgements}

The authors would like to thank Johannes Schmude, Marcus Freitag, Tom Theis, Maxim Lysak, Martin Mevissen, Naomi Simumba, and Héctor Maeso-García from IBM; Vincent Mai and Javad Bayazi from Hydro-Québec; Blazhe Gjorgiev from ETH Zurich; Jochen Stiasny and Olayiwola Arowolo from TU Delft for their valuable discussions and feedback.

\FloatBarrier
\bibliographystyle{IEEEtran}  
\bibliography{references}  

\clearpage

\appendix

\section{Comparison of Data Generation Methods and Libraries}
\label{app:comparison}
\begin{table}[H]
\centering
\renewcommand{\arraystretch}{1.2}
\label{tab:comparison}
\resizebox{\linewidth}{!}{%
\begin{tabular}{|P{3.6cm}|P{1.6cm}|P{4.5cm}|P{2cm}|P{3.6cm}|P{3.2cm}|P{2.2cm}|}
\hline
\textbf{Method / Library} & \textbf{Grid Scale} & \textbf{Load Perturbation} & \textbf{Topology Perturbation} & \textbf{Gen Setpoints Diversity} & \textbf{Net Param Perturbation} & \textbf{Open Source} \\
\hline
\textbf{\textcolor{blue}{gridfm-datakit}} &   &  & \checkmarkemoji & \checkmarkemoji & \xmark & \checkmarkemoji \\
\cite{gridfm-datakit2025} & Tested on up to 30,000 buses for PF and 10,000 for OPF  & Combined grid-level and bus-level scaling using real-world aggregated load profile & N-k for any k (line, transformer, generator) & Solved via OPF with permuted costs & Resistance and reactance scaled uniformly by a factor in \(( \max(0, 1 - \sigma), 1 + \sigma )\) where \(\sigma\) is a hyper-parameter & Code (Generated data to be made available by February 2025) \\
\hline
\textbf{OPFData} &   &  & \checkmarkemoji & \xmark & \xmark & \checkmarkemoji \\
\cite{lovett2024opfdatalargescaledatasetsac, piloto2024canosfastscalableneural} & Up to 13,659 buses & Uniform bus-wise scaling from [0.8, 1.2] & N-1 (line, transformer, generator) & Solved via OPF with fixed costs & Not applied & 600k samples x 10 grids \\
\hline
\textbf{OPFLearn} &   &  & \xmark & \xmark & \xmark & \checkmarkemoji \\
\cite{opflearn} & Up to 118 buses & Uniform sampling from convex feasible set (compute expensive) & No topology variation & Solved via OPF with fixed costs & Not applied & Julia package + 10k samples x 5 grids \\
\hline
\textbf{PF$\Delta$} &   & & \checkmarkemoji & \checkmarkemoji & \xmark & \checkmarkemoji \\
\cite{pfdelta} & Up to 2000 buses &  Uniform sampling from convex feasible set (compute expensive) & N-1 (line, transformer, generator) & Solved via OPF with permuted costs & Not applied & Code +  159,600  samples x 6 grids  (except for
GOC-2000, for which 61,800 samples are available) \\
\hline
\textbf{PGLearn} &   &  & \checkmarkemoji & \xmark & \xmark & \checkmarkemoji \\
\cite{pglearn} & Up to 24,000 buses & Combined grid-level and bus-level scaling & N-1 (line) & Solved via OPF with fixed costs & Not applied & Code + 10,000,000 samples over 14 grids. (about 500,000 samples per grid for small grids and 80,000 for grids > 6,000 buses) \\
\hline
\textbf{PowerGraph} &   &  & \xmark & \xmark & \xmark & \checkmarkemoji \\
\cite{varbella2024powergraph} & Up to 118 buses & Grid-level scaling using real-world aggregated load profile & No topology variation & Solved via OPF with fixed costs & Not applied & Data: 34,944 samples x 4 grids \\
\hline
\textbf{PowerFlowNet} &   &  & \xmark & \checkmarkemoji & \checkmarkemoji & \checkmarkemoji \\
\cite{LIN2024110112} & Up to 6470 buses & Normally perturbed loads (10\% std) & No topology variation & Sampled: voltage $\in$ [1.00, 1.05] p.u., active power sampled from normal distribution (10\% std) & Resistance and reactance scaled uniformly between 80\% and 120\% of their initial value & Code + 30k samples x 3 grids \\
\hline
\textbf{TypedGNN} &   &  & \checkmarkemoji & \checkmarkemoji & \checkmarkemoji & \xmark \\
\cite{LOPEZGARCIA2023105567} & Up to 118 buses & Uniform bus-wise scaling from [0.5, 1.5] & N-1 (line outage) & Sampled: voltage $\in$ [0.9, 1.1] p.u., active power sampled $\in$ [25\%, 75\%] of allowed range & Resistance, reactance, susceptance, and tap ratios scaled uniformly between 90\% and 110\% of their initial value & Not public \\
\hline
\textbf{Graph Neural Solver} &   &  & \xmark & \checkmarkemoji & \checkmarkemoji & \xmark \\
\cite{DONON2020106547, BalthazarDonon2022DeepApplications} & Up to 118 buses &  Uniform bus-wise scaling from [0.5, 1.5] & No topology variation & Sampled: voltage $\in$ [0.95, 1.05] p.u., active power sampled $\in$ [25\%, 75\%] of allowed range & Resistance, reactance, and susceptance scaled uniformly between 90\% and 110\% of their initial value; taps sampled $\in$ [0.8, 1.2]; shift sampled $\in$ [–0.2, 0.2] rad & Not public \\
\hline
\end{tabular}
}
\end{table}

\section{Details on Data Generation Runtime}
\label{app:data_gen}

Data generation time corresponds to processing 10{,}000 load scenarios and creating 20 power flow topology variants for each, yielding a total of 200{,}000 samples. The computations were performed on a cluster equipped with Intel(R) Xeon(R) Gold 6258R CPUs @ 2.70\,GHz and AMD EPYC 7763 CPUs @ 2.45\,GHz. For most datasets, 20 cores were used, except for OPF GOC 2k (60 cores) and GOC 10k (100 cores). 32\,GB of RAM sufficed for all datasets except GOC 10k, which required 256\,GB. The default linear solver provided with Ipopt.jl was used; however,~\cite{powermodels} reports that runtime can be reduced by 2--6$\times$ using the HSL ma57 solver.  

When generating PF data, runtime is primarily driven by the number of load scenarios: an OPF is solved once per load scenario on the base topology, while only PF (which is significantly faster than OPF) is solved for each of the 20 topology variants. Consequently, increasing the number of topology variants per load scenario is an efficient way to obtain more samples. In OPF mode, data generation takes longer since OPF is solved for each topology variant.

\section{Computation of the Mean Normalized Shannon Entropy}

\label{app:shannon}

We quantify dataset diversity using the \emph{mean normalized Shannon entropy}. Motivations for using such a metric are discussed in~\cite{hedgeopf2025}. Standard deviation may overestimate variability when a feature takes only a few discrete values (e.g., \(P_g\) switching between zero and its upper bound). In contrast, Shannon entropy measures the \emph{distributional uncertainty} of a feature: it is low when samples concentrate on a small portion of the admissible domain and increases only when the empirical distribution spreads across multiple well-populated regions.

\paragraph{Histogram Domains.}

Entropy is computed from a discretized approximation of the empirical distribution at each bus. We construct fixed-range histograms using the interval \((-\pi,\pi]\) for \(V_a\), and empirical per-bus minima and maxima (different for PF and OPF) across datasets for \(V_m, P_d, Q_d, P_g, Q_g\) (operational limits for these cannot be used since they can be violated in the case of PF). We use consistent domains across datasets to ensure that entropy values are comparable and not driven by binning inconsistencies.

\paragraph{Per-Bus Shannon Entropy.}

A 100-bin histogram over the specified domain yields empirical probabilities \(\{p_i\}_{i=1}^{100}\) over the bins. The Shannon entropy at that bus is
\[
H = - \sum_{i=1}^{100} p_i \log_2 p_i,
\]
with \(0 \log 0 = 0\).

\paragraph{Averaging and Normalization.}

Entropies are averaged across buses and normalized by the maximum value \(\log_2(100)\), yielding the \emph{mean normalized Shannon entropy},
\[
\tilde{H} = \frac{1}{\log_2(100)} \cdot \frac{1}{B} \sum_{b=1}^{B} H_b \in [0,1].
\]
Values near zero indicate that the dataset exhibits little variability, whereas values approaching one correspond to broad and nearly uniform exploration of the feature’s admissible range.

\clearpage

\section{\datakit Outputs}
\label{app:outputs}

\begin{figure}[H]
    \centering
    \begin{subfigure}[t]{0.2\textwidth}
        \centering
        \includegraphics[width=\linewidth]{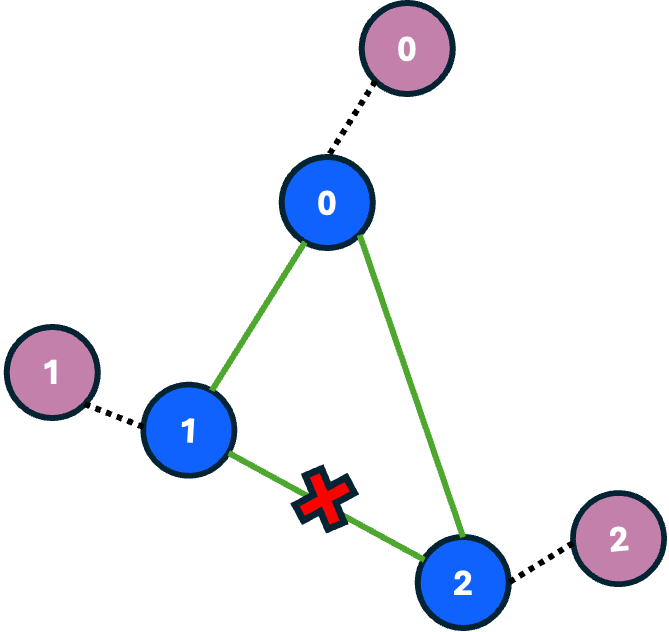}
        \caption{}
    \end{subfigure}\hfill
    \begin{subfigure}[t]{0.7\textwidth}
        \centering
        \includegraphics[width=\linewidth]{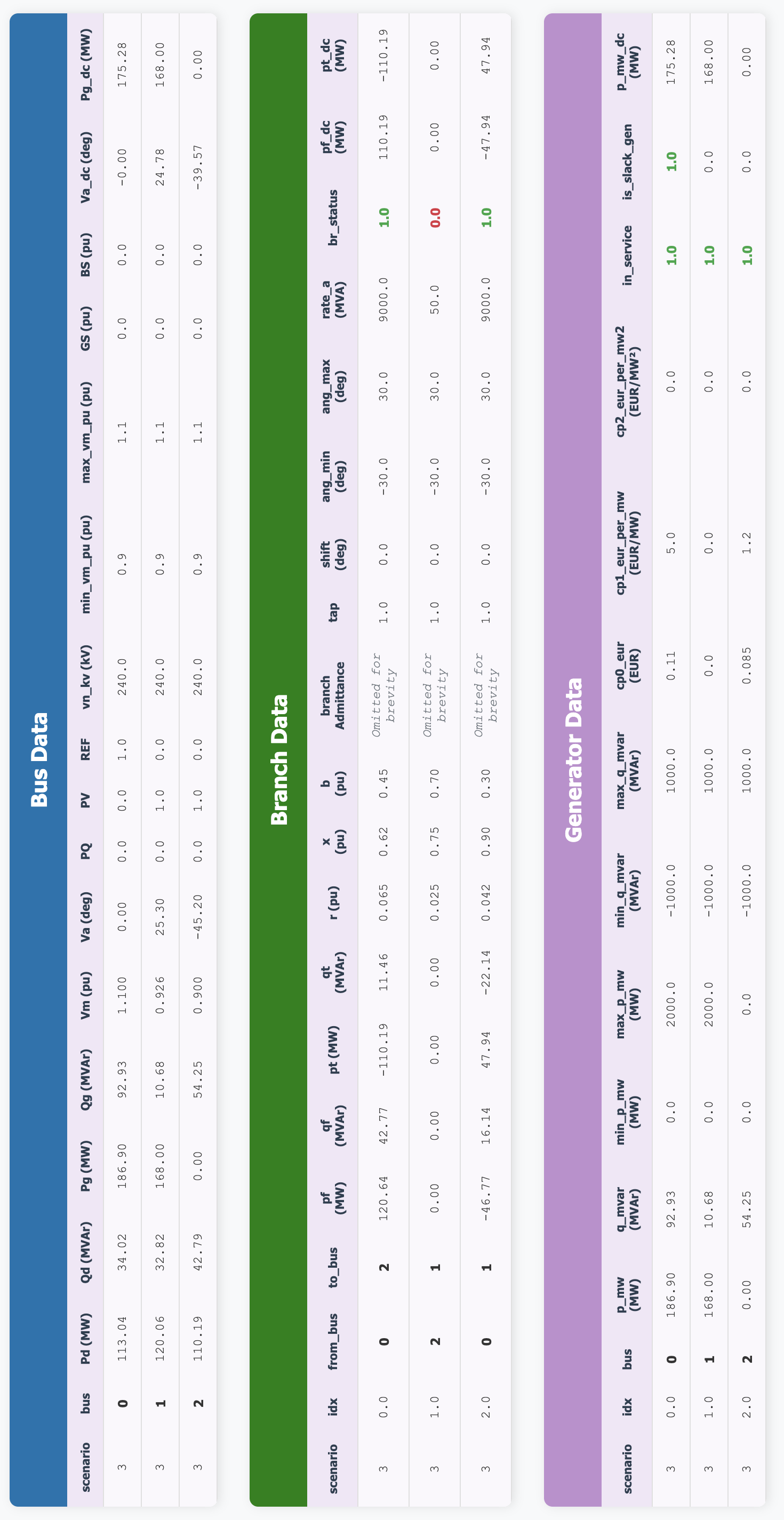}
        \caption{}
    \end{subfigure}
    \caption{\datakit output datasets: (a) network topology showing buses, branches, and generators for the first scenario; (b) corresponding data for the same scenario.}
    \label{fig:outputs}
\end{figure}

\begin{figure}[H]
\centering 
\includegraphics[width=\linewidth]{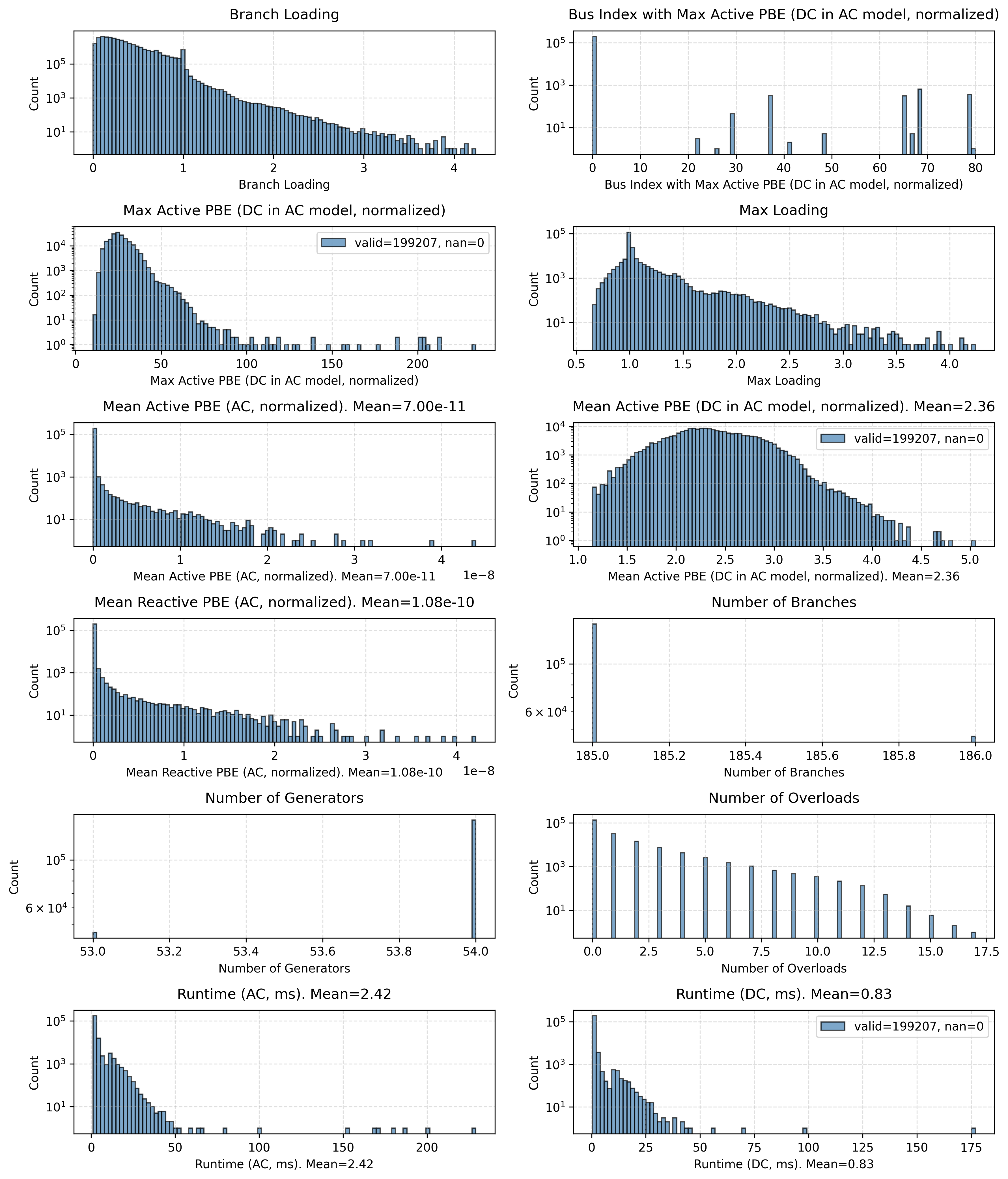}
\caption{Example output of the \texttt{gridfm\_datakit stats} command on PF data generated for IEEE 118.}
\label{fig:stats} 
\end{figure}

\section{Violin plots}
\label{app:violin}

\begin{figure}[t]
    \centering

    \begin{subfigure}[t]{0.48\textwidth}
        \centering
        \includegraphics[width=\linewidth]{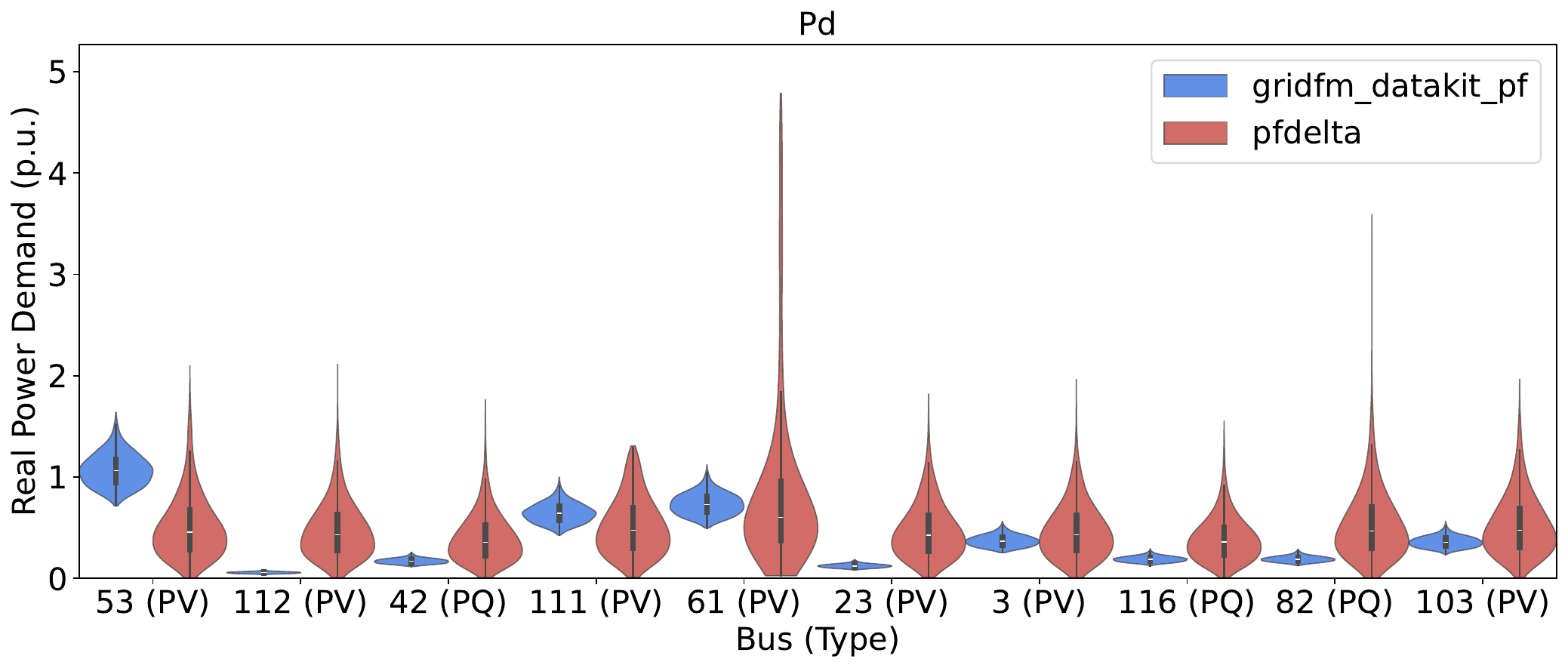}
        \caption{PF: $P_d$}
    \end{subfigure}\hfill
    \begin{subfigure}[t]{0.48\textwidth}
        \centering
        \includegraphics[width=\linewidth]{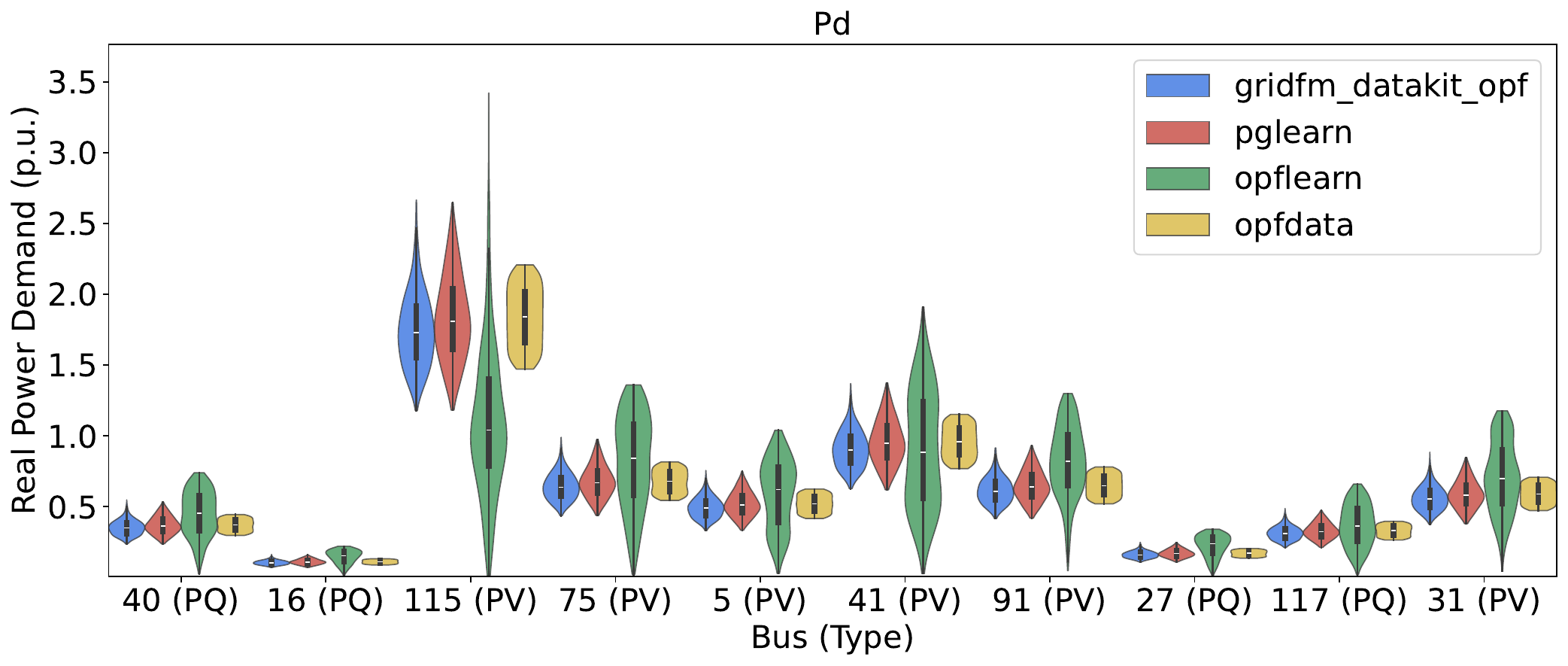}
        \caption{OPF: $P_d$}
    \end{subfigure}

    \vspace{0.3cm}

    \begin{subfigure}[t]{0.48\textwidth}
        \centering
        \includegraphics[width=\linewidth]{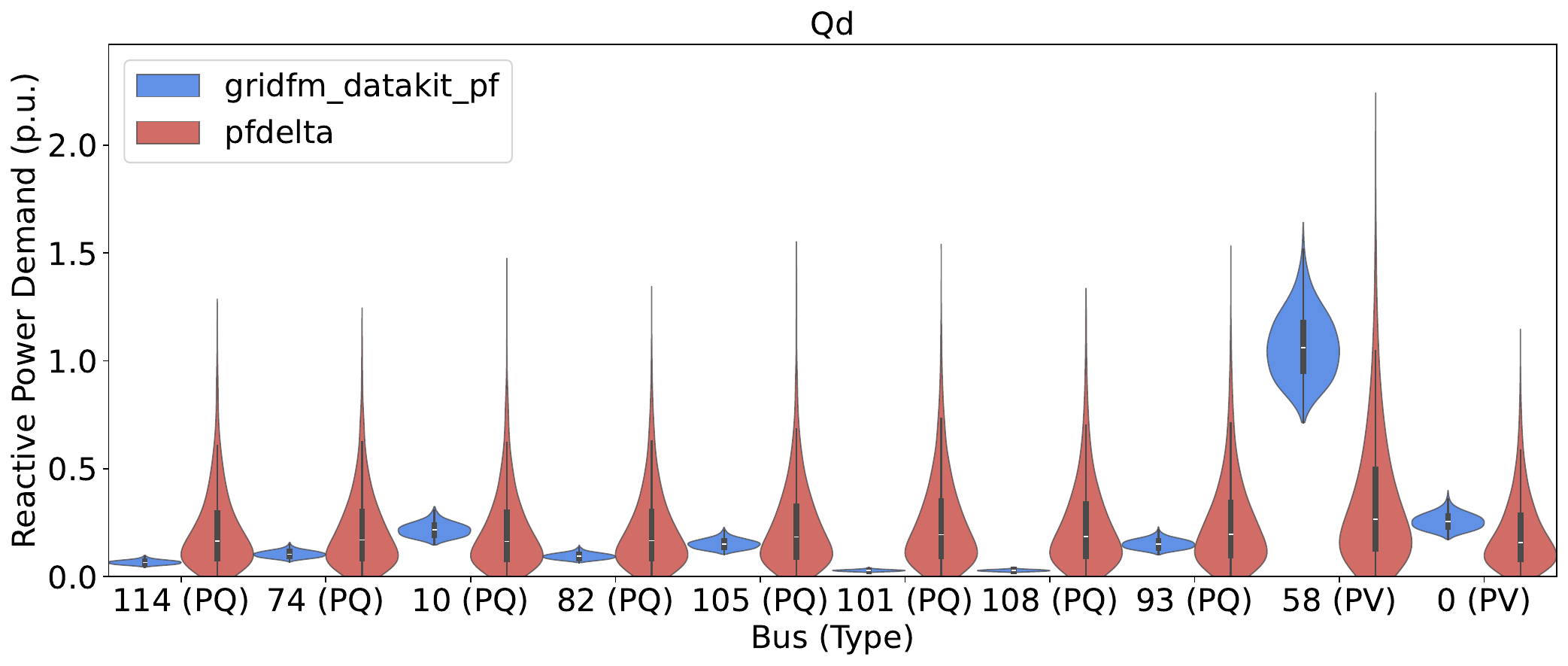}
        \caption{PF: $Q_d$}
    \end{subfigure}\hfill
    \begin{subfigure}[t]{0.48\textwidth}
        \centering
        \includegraphics[width=\linewidth]{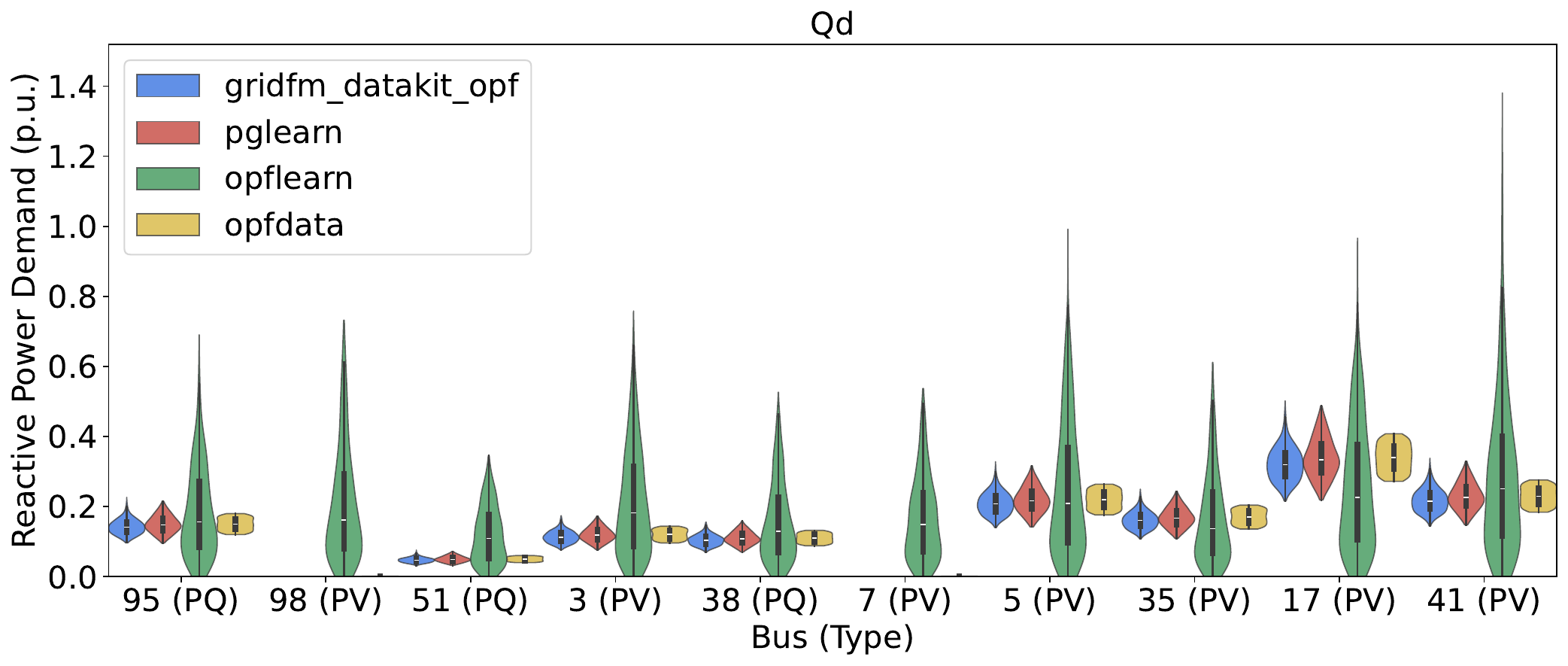}
        \caption{OPF: $Q_d$}
    \end{subfigure}

    \vspace{0.3cm}

    \begin{subfigure}[t]{0.48\textwidth}
        \centering
        \includegraphics[width=\linewidth]{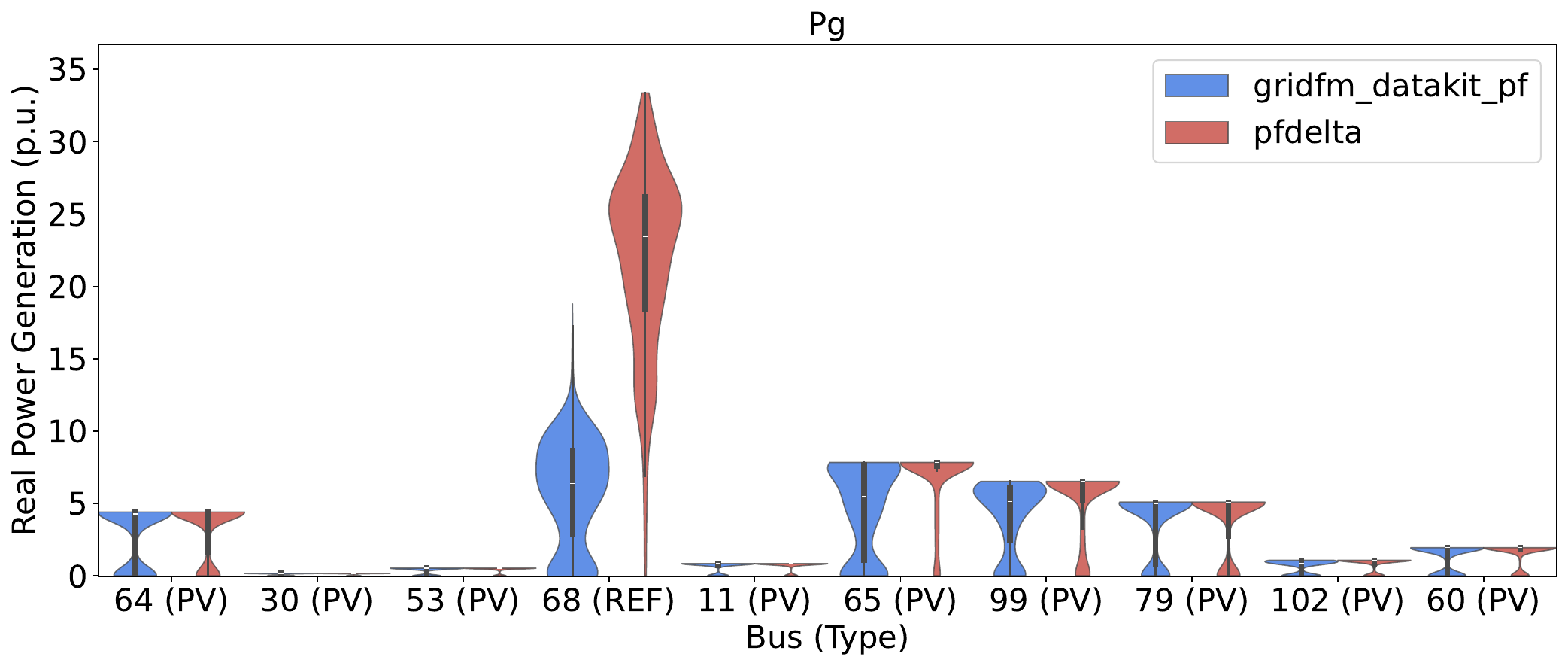}
        \caption{PF: $P_g$}
    \end{subfigure}\hfill
    \begin{subfigure}[t]{0.48\textwidth}
        \centering
        \includegraphics[width=\linewidth]{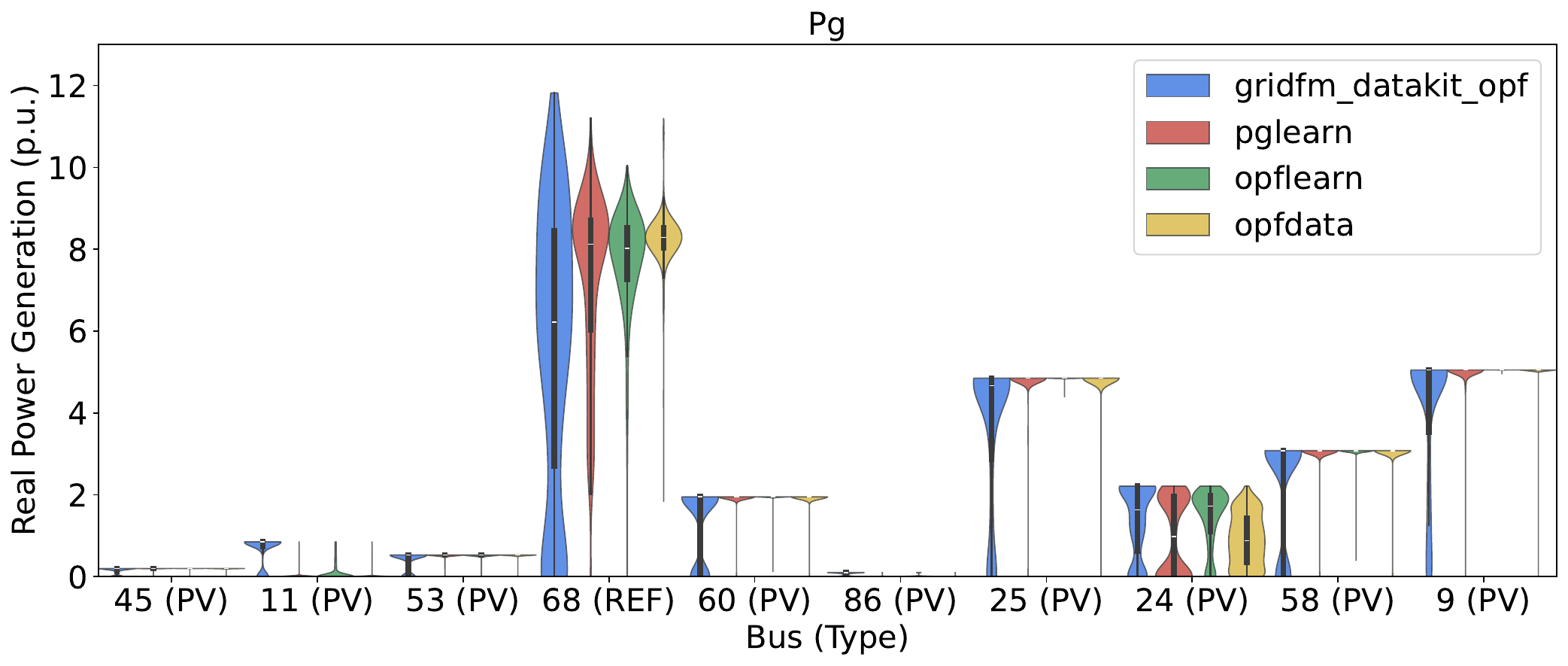}
        \caption{OPF: $P_g$}
    \end{subfigure}

    \vspace{0.3cm}

    \begin{subfigure}[t]{0.48\textwidth}
        \centering
        \includegraphics[width=\linewidth]{figures/comparison_plots/Qg_violin_pf.pdf}
        \caption{PF: $Q_g$}
    \end{subfigure}\hfill
    \begin{subfigure}[t]{0.48\textwidth}
        \centering
        \includegraphics[width=\linewidth]{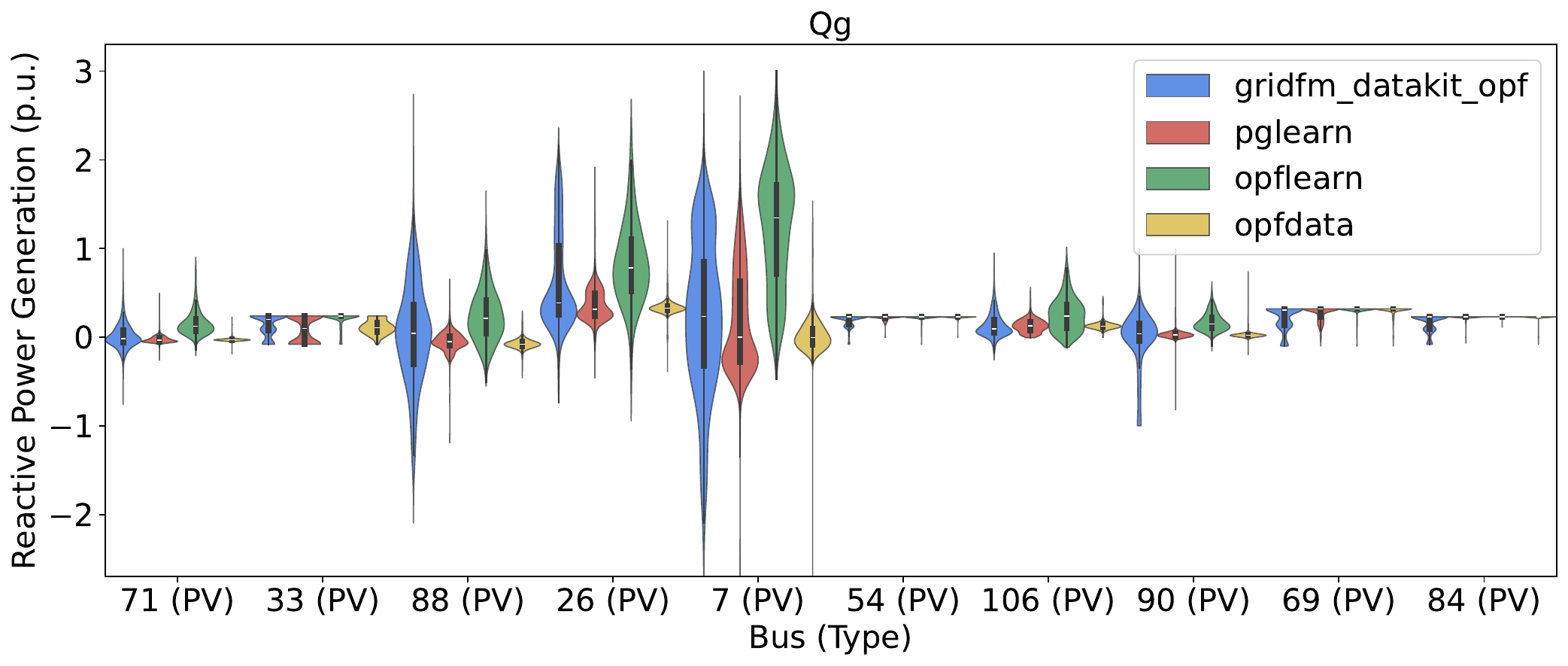}
        \caption{OPF: $Q_g$}
    \end{subfigure}

    \vspace{0.3cm}

    \begin{subfigure}[t]{0.48\textwidth}
        \centering
        \includegraphics[width=\linewidth]{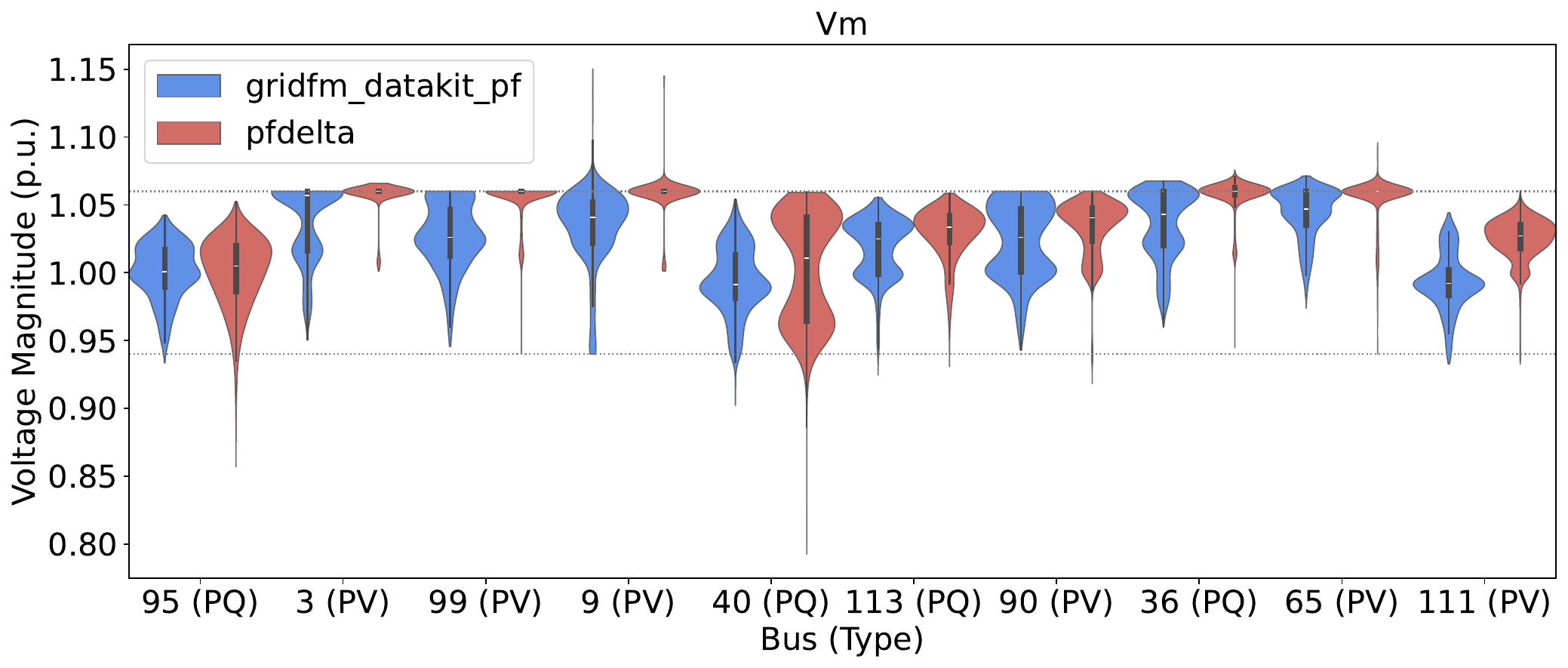}
        \caption{PF: $V_m$}
    \end{subfigure}\hfill
    \begin{subfigure}[t]{0.48\textwidth}
        \centering
        \includegraphics[width=\linewidth]{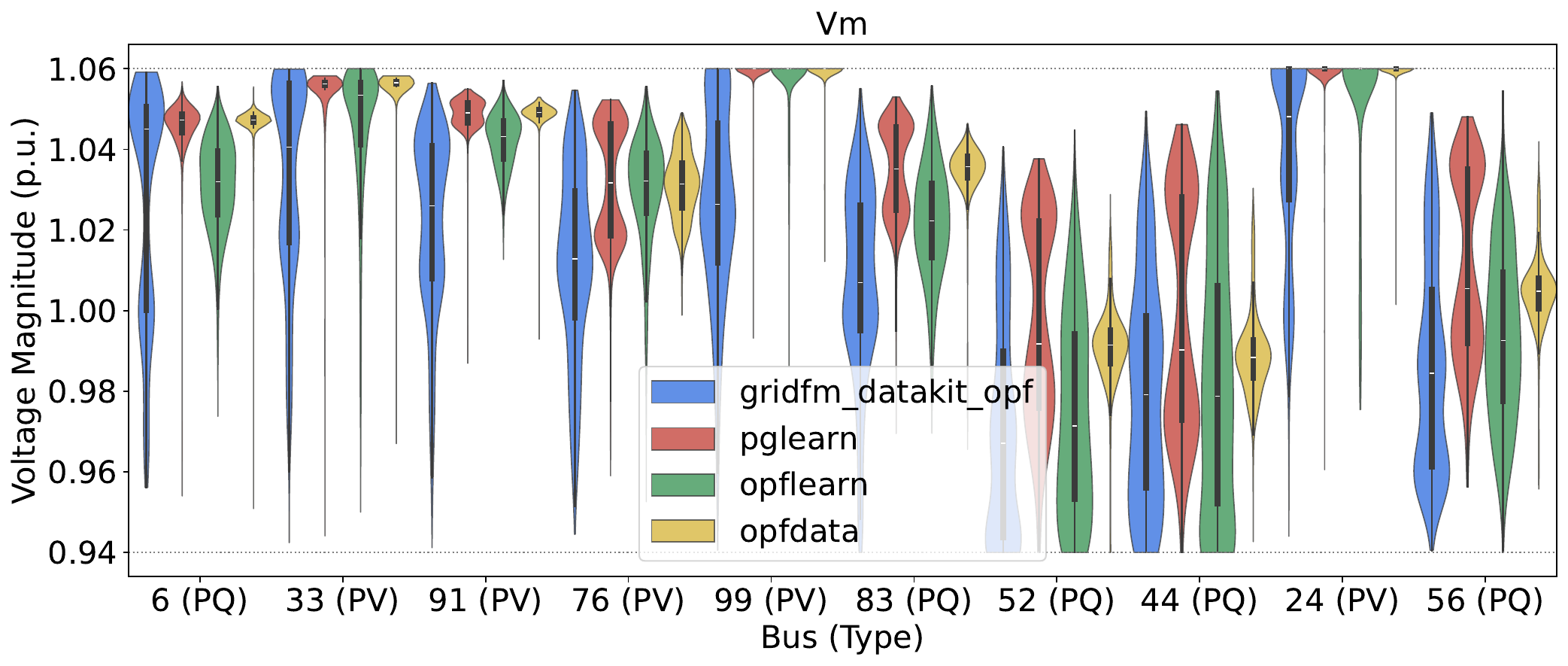}
        \caption{OPF: $V_m$}
    \end{subfigure}

    \vspace{0.3cm}

    \begin{subfigure}[t]{0.48\textwidth}
        \centering
        \includegraphics[width=\linewidth]{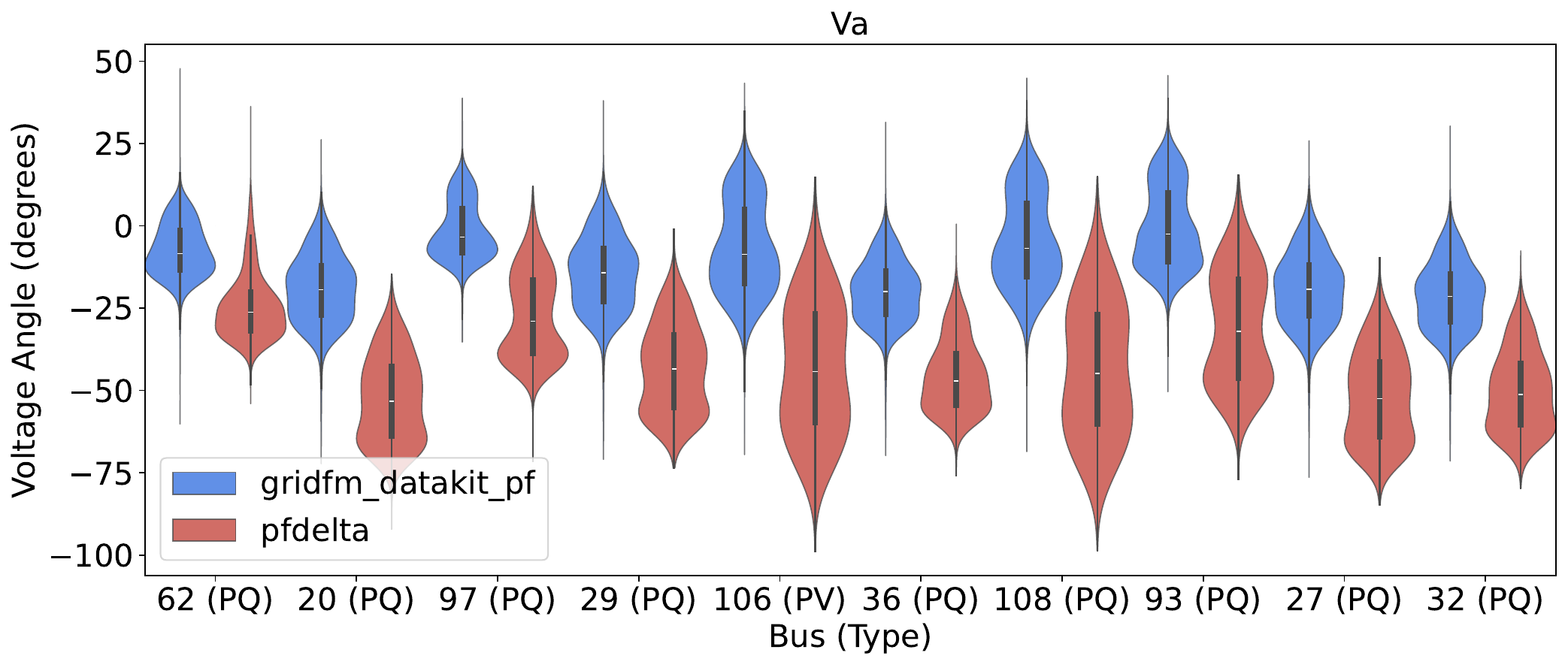}
        \caption{PF: $V_a$}
    \end{subfigure}\hfill
    \begin{subfigure}[t]{0.48\textwidth}
        \centering
        \includegraphics[width=\linewidth]{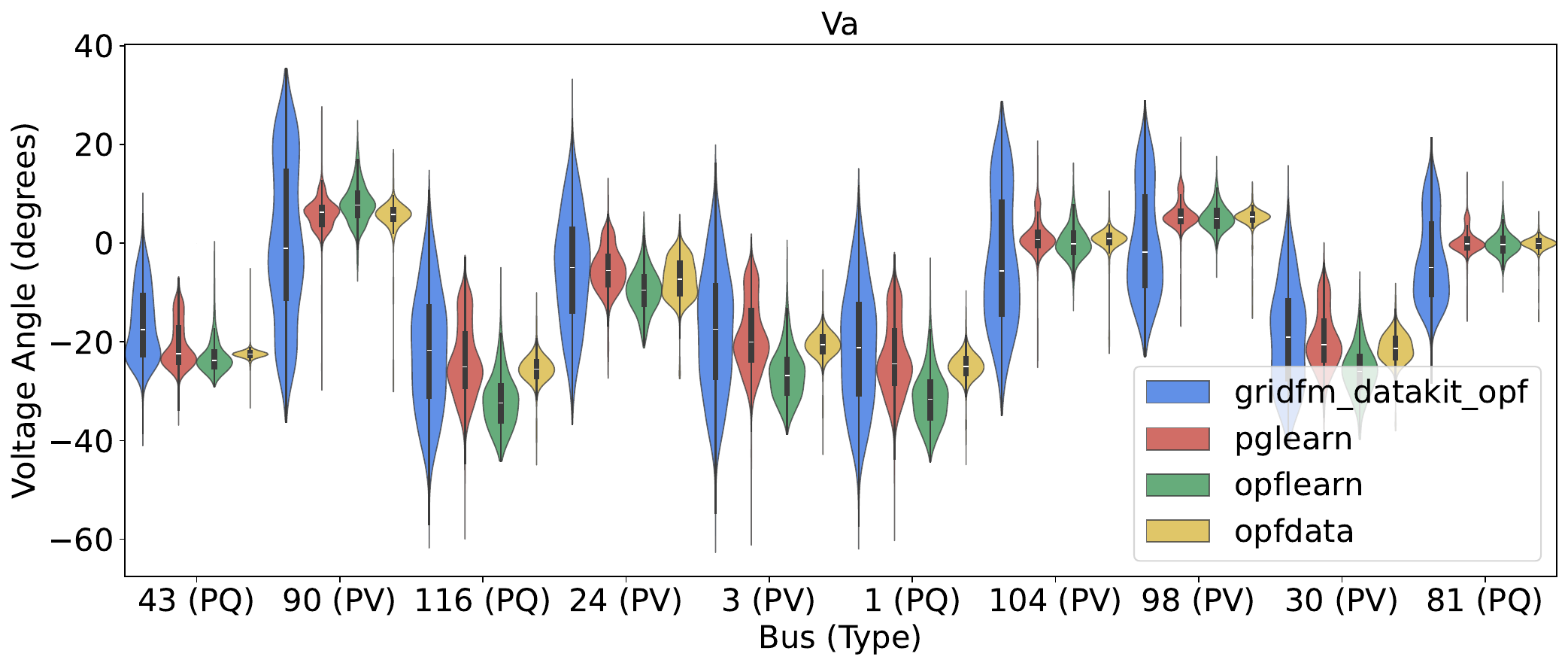}
        \caption{OPF: $V_a$}
    \end{subfigure}

    \caption{Feature distributions for Power Flow (PF) and Optimal Power Flow (OPF) datasets. Left column: PF; Right column: OPF. Features shown are $P_d$, $Q_d$, $P_g$, $Q_g$, $V_m$, and $V_a$.}
    \label{fig:feature_violin_type}
\end{figure}

\end{document}